\documentclass{article}
\usepackage{iclr2027_conference,times}
\usepackage{amsmath,amssymb,amsthm,booktabs,graphicx,tabularx,array}
\usepackage{hyperref,url}
\usepackage{letterspace}
\usepackage[above]{placeins}
\usepackage{needspace}
\usepackage{etoolbox}
\makeatletter
\patchcmd{\@maketitle}
  {\begin{tabular}[t]{l}\bf\rule{\z@}{24pt}\@author\end{tabular}}
  {\makebox[\textwidth][c]{\begin{tabular}[t]{c}\bf\rule{\z@}{24pt}\@author\end{tabular}}}
  {}
  {\PackageError{arxiv-layout}{Unable to center the author block}{Check the title template.}}
\patchcmd{\@maketitle}
  {\vskip 0.3in minus 0.1in}
  {\vskip 0.12in minus 0.03in}
  {}
  {\PackageError{arxiv-layout}{Unable to adjust author spacing}{Check the title template.}}
\makeatother
\newcommand{\posvariant}{Positional}

\newtheorem{proposition}{Proposition}
\title{\textls[-40]{Rethinking Memorization Mitigation in\\
Diffusion Models: Reinforcing Text Conditioning}}
\author{%
\textbf{Hyungjun Joo}$^{1}$ \quad \textbf{Sehwan Kim}$^{2}$ \quad \textbf{Hyeonggeun Han}$^{2}$ \quad \textbf{Sangwoo Hong}$^{3}$ \quad \textbf{Jungwoo Lee}$^{2}$\\[0.2em]
\normalfont $^{1}$Samsung Electronics \quad $^{2}$Seoul National University \quad $^{3}$Konkuk University\\[0.3em]
{\small\normalfont\texttt{hyjun.joo@samsung.com, sehwankim@snu.ac.kr, hygnhan@snu.ac.kr}}\\
{\small\normalfont\texttt{swhong06@konkuk.ac.kr, junglee@snu.ac.kr}}
}
\iclrfinalcopy
\hypersetup{
  hidelinks,
  pdftitle={Rethinking Memorization Mitigation in Diffusion Models: Reinforcing Text Conditioning},
  pdfauthor={Hyungjun Joo, Sehwan Kim, Hyeonggeun Han, Sangwoo Hong, Jungwoo Lee},
  pdfdisplaydoctitle=true
}
\begin{document}
\maketitle
% The template's camera-ready switch exposes authors; clear its publication header.
\fancyhead{}
\renewcommand{\headrulewidth}{0pt}
\begin{abstract}
Text-to-image diffusion models have achieved remarkable progress in image synthesis, yet can exhibit memorization by closely reproducing individual training examples. Effective mitigation must preserve useful prompt information to guide alternative depictions. We introduce a training-free method that redistributes cross-attention with Gaussian smoothing before reinforcing content-token contributions and attenuating padding contributions, without additional denoiser evaluations. With this intervention, stronger content conditioning can improve prompt alignment at comparable training-image similarity. A local analysis identifies when reinforcement preserves shared value information while redistribution reduces localized attention mass. On Stable Diffusion v1.4 and v2.0, all evaluated smoothing widths lie on the empirical Pareto frontiers for training-image similarity versus both prompt alignment and image preference. A configuration selected on Stable Diffusion reduces template reproduction in DeepFloyd IF without further tuning. These findings support jointly controlling conditioning allocation and strength to generate prompt-consistent alternatives.
\end{abstract}

\section{Introduction}
\label{sec:introduction}

\begin{figure}[t]
\centering
\includegraphics[width=\linewidth]{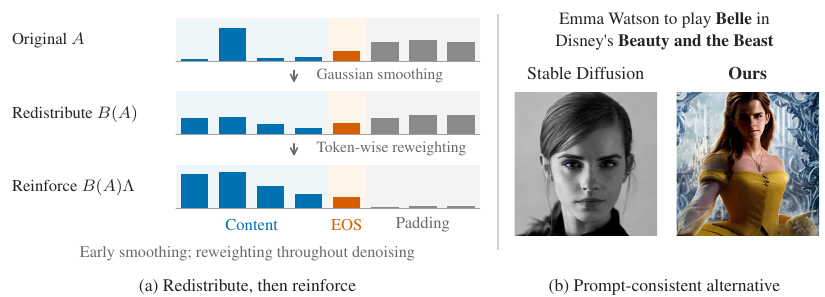}
\par\vspace{-6pt}
\caption{\textbf{Reinforce a redistributed conditioning pattern.} (a) An illustrative attention row shows Gaussian redistribution followed by content reinforcement and padding attenuation, on a shared scale. (b) Recorded outputs illustrate an alternative depiction that retains the prompt context.}
\label{fig:teaser}
\vspace{-6pt}
\end{figure}

Text-to-image diffusion models enable high-quality image synthesis from natural-language descriptions \citep{DDPM,SDE1,SD,Dall-E,Imagen}. Yet these models can exhibit \emph{memorization}: certain prompts elicit close reproductions of individual training images \citep{Memorization_usenix,Memorization_forgery,Memorization_understanding,Memorization_500prompt}. Such reproduction can preserve a training image's distinctive composition and visual details, beyond the semantic resemblance expected from depicting the same subject. Mitigating this behavior requires more than moving the generated image away from a training reference: the model must still produce a useful alternative that follows the prompt. A prompt generally permits many valid depictions, and its objects, attributes, and context provide information for constructing them. Figure~\ref{fig:teaser} illustrates this goal: the film context in the Emma Watson/Belle prompt supports an alternative depiction of the subject.

Several inference-time approaches mitigate memorization by weakening or avoiding conditional responses associated with reproduction. Prompt-embedding adjustment reduces the magnitude of text-conditional guidance \citep{Memorization_detecting}, while initial-noise adjustment seeks starting points with weaker conditional responses and earlier escape from memorization basins \citep{AIN}. Other approaches postpone or reverse guidance \citep{Memorization_basin}, or reduce attention to prompt words \citep{Memorization_unveiling}. We instead make \emph{content reinforcement} an active part of mitigation: strengthening useful prompt information to guide alternative depictions.

We find that content reinforcement can improve prompt alignment at comparable training-image similarity. This improvement occurs when content gain is increased with redistribution and padding attenuation held fixed, showing that reinforcement can support useful conditioning during mitigation. We implement this approach as a training-free cross-attention intervention that combines content amplification with Gaussian redistribution and selective padding attenuation. Early smoothing changes the allocation of attention across token positions; amplification then strengthens the redistributed content contributions. The intervention retains the original prompt and frozen model and requires no additional denoiser evaluations.

Redistribution improves the suppression--quality balance of this reinforcement strategy. At fixed token gains in our main configuration, Gaussian redistribution recovers prompt alignment and image preference over reweighting alone while retaining most of the suppression. A local analysis complements these generation-level results: under value-coherence conditions, an admissible gain preserves shared feature information while reducing the attention component's mass on its original token set. Together, these results explain how controlling attention allocation supports reinforcement during mitigation.

On two Stable Diffusion benchmarks, all evaluated smoothing widths lie on the empirical Pareto frontiers for training-image similarity versus prompt alignment and image preference. A direction--norm control shows that suppression persists after restoring guidance magnitude. Our configuration selected on Stable Diffusion also transfers to DeepFloyd IF without IF-specific tuning. These findings support content reinforcement as a means of generating useful alternatives while mitigating reproduction.

\section{Related work}
\label{sec:related_work}

\paragraph{Memorization in diffusion models.}
Training-image extraction and replication studies show that diffusion models can reproduce individual training examples \citep{Memorization_usenix,Memorization_forgery}. Data duplication and image--caption associations influence this behavior \citep{Memorization_understanding,Memorization_forgery,chenchen}. Mechanistic analyses locate memorization-related neurons in cross-attention layers \citep{Memorization_nemo}, while geometric analyses explain memorization through the relationship between learned and data manifolds \citep{Memorization_LID}. Reproduction also takes different forms: \citet{Memorization_500prompt} documents both image and template reproduction, and Bright Ending \citep{BrightEnding} examines local memorization in which only part of an image is copied. These perspectives provide complementary accounts of what is reproduced and how memorization is represented in the model.

\paragraph{Inference-time mitigation.}
Existing methods intervene at different points in generation. Random prompt perturbations disrupt the conditioning associated with reproduction \citep{Memorization_understanding}, while gradient-based prompt adjustment optimizes text embeddings to reduce the magnitude of text-conditional noise predictions \citep{Memorization_detecting}. Guidance-based methods delay or reverse classifier-free guidance to escape memorization basins \citep{Memorization_basin}; Anti-Memorization Guidance combines guidance terms targeting image duplication, caption duplication, and prompt specificity \citep{chenchen}. Initial-noise adjustment searches for starting points that permit earlier escape from memorization basins \citep{AIN}. Bright Ending \citep{BrightEnding} localizes copied regions using EOS-token attention at the final denoising step and uses the resulting spatial mask to focus detection and prompt optimization on those regions. For direct attention modification, \citet{Memorization_unveiling} combine summary-token masking with an increase in the start-of-sequence (SOS) token's logit before softmax; the latter shifts attention toward SOS and reduces the attention weights assigned to prompt words. Our method instead redistributes attention across token positions before reinforcing content contributions, preserving the original prompt and initial noise without input optimization.

\paragraph{Attention control for semantic fidelity.}
Cross-attention control also supports the expression of prompt semantics. Prompt-to-Prompt \citep{Cross_prompttoprompt} reuses attention maps to preserve spatial layout during text-based image editing. Structured Diffusion \citep{Cross_structure_diffusion} incorporates linguistic structure into cross-attention guidance to improve compositional generation. SynGen \citep{Cross_syngen} uses syntactic relations to align the attention maps of entities and their modifiers, improving attribute binding. Attend-and-Excite \citep{Cross_excite} addresses neglected subjects by smoothing their spatial attention maps and iteratively optimizing the latent to strengthen their activations. We share the goal of making generated images more faithfully reflect prompt content. Our intervention redistributes attention along token positions and directly amplifies the resulting content contributions, coupling this semantic objective with memorization mitigation rather than optimizing a subject-specific spatial attention objective.

\section{Reinforcing redistributed cross-attention}
\label{sec:method}

We reinforce prompt content to support useful alternatives during mitigation, while controlling its allocation across token positions (Figure~\ref{fig:teaser}). We first redistribute attention, then reinforce content within the resulting pattern and selectively attenuate padding contributions. This order makes amplification act on a new mixture of token values. We formalize the operator, analyze reinforcement with localized dilution, and adapt it to DeepFloyd IF.

\subsection{Redistribute, reinforce, and preserve context}
At sampling step $k\in\{0,\ldots,T-1\}$, cross-attention computes
\begin{equation}
 A_k=\operatorname{softmax}(Q_kK_k^\top/\sqrt{d}),\qquad O_k=A_kV_k.
\end{equation}
We intervene on each attention row and head in the conditional branch. Under SD's CLIP tokenization, let $s$ denote the start-token position and $e$ the first EOS position. The content span is $\mathcal C=\{s+1,\ldots,e-1\}$ and the padding span comprises positions after the first EOS, $\mathcal P=\{e+1,\ldots,L-1\}$. Content denotes prompt-token positions, including subwords, rather than a separately inferred set of important words.

\paragraph{Redistribution: control the extent of mixing.}
A concentrated attention row repeatedly emphasizes a small part of the available value information. We first change this allocation by applying a normalized, truncated Gaussian kernel along token positions $1,\ldots,L-1$, using width $\sigma$ and reflection padding. The start-token column is unchanged. Denote this operator by $B_k$. Gaussian weights provide a distance-dependent redistribution rule: nearby positions exchange more attention, and $\sigma$ controls how widely each contribution spreads. This supplies a simple, tunable continuum from narrow mixing to broad redistribution without selecting individual subjects or optimizing a prompt-specific attention target.

The positional kernel defines the extent of mixing; the value vectors determine what information that mixing retains. Local consistency of value information, formalized in Section~\ref{sec:theory}, governs this retention. Wider kernels dilute localized contributions more strongly but also mix across more varied values. We apply smoothing during an early denoising window to establish the redistributed conditioning pattern, then let $B_k$ become the identity for the remaining steps. We use the first 5 of 50 steps for SD v1.4 and the first 10 for SD v2.0.

\paragraph{Reinforcement: support content after redistribution.}
We reweight redistributed attention as
\begin{equation}
\lambda_j=\begin{cases}\lambda_w,&j\in\mathcal C,\\ \lambda_p,&j\in\mathcal P,\\1,&\text{otherwise,}\end{cases}
\qquad \lambda_w\geq1,\quad 0\leq\lambda_p\leq1.
\end{equation}
The content gain $\lambda_w$ strengthens the prompt information carried by the new mixture. Applying it after redistribution assigns gain according to the \emph{recipient} token: attention reaching content is reinforced, whereas attention reaching padding is attenuated. Reinforcement acts on the conditioning pattern created by smoothing, allowing useful content contributions to grow while the original concentration can remain diluted. We use a common content gain to retain all prompt-token positions as potential sources of an alternative depiction.

\paragraph{Selective suppression: retain a contextual anchor.}
The padding gain $\lambda_p$ controls contributions beyond the content span. Prior work shows that EOS representations carry contextual information useful for generation \citep{EOS_good}. Motivated by this observation, we preserve the first EOS while attenuating subsequent padding, and evaluate this boundary choice in Section~\ref{sec:eos}. EOS is exempt from multiplicative suppression, although smoothing can change its attention. This boundary complements reinforcement: explicit prompt tokens receive additional weight, the first EOS provides a contextual anchor, and later positions receive reduced weight.

Writing $\Lambda=\operatorname{diag}(\lambda_j)$, the complete intervention is
\begin{equation}
 \widehat A_k=(1-\alpha)A_k+\alpha B_k(A_k)\Lambda,\qquad
 \widehat O_k=\widehat A_kV_k.
 \label{eq:operator}
\end{equation}
We use $\lambda_w=2$ and $\lambda_p=0.1$. The coefficient $\alpha\in[0,1]$ sets the intervention strength, with $\alpha=1$ in the main baseline comparison. Both terms use the same attention logits. Rows retain their reweighted mass, so token gains can affect both the composition and magnitude of $\widehat O_k$. Weighting continues throughout sampling to support content conditioning after the early redistribution window. Token embeddings, model parameters, and the unconditional branch remain unchanged. The intervention adds attention transformations without increasing the number of denoiser evaluations.

\subsection{Reinforcement with localized dilution}
\label{sec:theory}
Redistribution changes the value mixture on which reinforcement acts. To make this explicit, consider one head during the smoothing window and write an attention row as a column vector $a$. Let $G$ be the linear blur operator, so that $B(a)=Ga$, and let the rows of $V$ contain the value vectors. Holding these vectors fixed as in Eq.~\eqref{eq:operator}, the transformed output satisfies the exact identity
\begin{equation}
 \widehat o^\top=a^\top\big[(1-\alpha)V+\alpha G^\top\Lambda V\big].
 \label{eq:mixture_view}
\end{equation}
Thus, the original attention weights act on a modified mixture of value vectors. Reinforcement amplifies recipients of redistribution, so it need not recover the original conditioning pattern. This distinction matters when reproduction depends on a few token embeddings \citep{Memorization_unveiling}.

\paragraph{An illustrative kernel calculation.}
Consider a component concentrated on one interior content token, with its entire Gaussian neighborhood inside the content span. For the implemented kernel at $\sigma=1$, the central coefficient is $g_0\approx0.399$. With full intervention ($\alpha=1$) and content gain $\lambda_w=2$, approximately $2g_0=0.798$ of the component's original mass returns to that position, even though its total mass across content doubles. Reinforcement can therefore increase overall content weight while reducing the weight retained at the original location.

The remaining question is what information survives this redistribution. If neighboring value vectors carry a common feature, the dispersed component can retain that feature even as its token allocation changes. To interpret the retained mass, consider a fixed additive value vector present at every position in the original token set and absent elsewhere. Its contribution from the attention component is proportional to the mass remaining on those positions. The two bounds below thus describe how a shared feature can be preserved while a localized contribution is attenuated. The analysis separates variation in the shared feature from the fraction of mass returned to the original set.

\paragraph{Local value coherence.}
Let $a\geq0$ be an attention component supported on a token set $\mathcal T\subset\mathcal C$, with mass $c=\sum_i a_i>0$, and write $o(a)=\sum_i a_i v_i$. For the local analysis, its kernel neighborhood lies inside the content span and avoids the reflected endpoints. Fix a unit feature direction $u$ and let $s_i=u^\top v_i$. Assume $s_i\geq s_*>0$ on $\mathcal T$ and $|s_j-s_i|\leq\ell|j-i|$ for its kernel neighbors. This condition bounds variation in a shared feature across neighboring value vectors.

For the implemented normalized kernel $g_r$, extended by zero outside its radius, define
\begin{equation}
 \mu_\sigma=\sum_r |r|g_r,\qquad
 \kappa_{\mathcal T}=\max_{i\in\mathcal T}\sum_{j\in\mathcal T}g_{j-i},\qquad
 \varepsilon=\frac{\ell\mu_\sigma}{s_*}.
 \label{eq:local_quantities}
\end{equation}
Here $\mu_\sigma$ measures transport distance, $\kappa_{\mathcal T}$ bounds the fraction returned to $\mathcal T$, and $\varepsilon$ bounds relative variation in the preserved feature. In particular, $\kappa_{\mathcal T}\leq\min\{1,|\mathcal T|g_0\}$.

\begin{proposition}[Reinforcement with localized dilution]
\label{prop:reinforcement}
Under the local value-coherence conditions, let $0<\alpha\leq1$, $S(a)=u^\top o(a)$, and $M_{\mathcal T}(a)=\sum_{j\in\mathcal T}a_j$. The component transformed by Eq.~\eqref{eq:operator} satisfies
\begin{align}
 S(\widehat a)&\geq[1-\alpha+\alpha\lambda_w(1-\varepsilon)]S(a),\label{eq:semantic_bound}\\
 M_{\mathcal T}(\widehat a)&\leq[1-\alpha+\alpha\lambda_w\kappa_{\mathcal T}]c.\label{eq:localized_bound}
\end{align}
If $\varepsilon+\kappa_{\mathcal T}<1$, every gain
\begin{equation}
 \frac{1}{1-\varepsilon}\leq\lambda_w<\frac{1}{\kappa_{\mathcal T}}
 \label{eq:gain_window}
\end{equation}
preserves or increases $S(a)$ while strictly reducing the component's mass assigned to $\mathcal T$.
\end{proposition}

The two sides of Eq.~\eqref{eq:gain_window} have complementary roles. The lower bound supplies enough gain to compensate for variation in the shared feature; the upper bound prevents amplification from restoring the original localized mass. For the singleton example above, the upper bound is approximately $2.51$, so gain 2 still dilutes the original position and preserves the projected feature whenever $\varepsilon\leq0.5$.

The gain window also clarifies the distinct roles of redistribution and reinforcement. When $\varepsilon+\kappa_{\mathcal T}<1$, the ratio of localized-mass retention to shared-feature retention at full mix is bounded by $\kappa_{\mathcal T}/(1-\varepsilon)<1$, independently of the common gain. Redistribution creates this relative advantage for the shared feature; reinforcement within the admissible window converts it into absolute feature preservation while keeping localized mass below its original level.

Appendix~\ref{app:theory} gives the proof and a full-row extension covering background attention, reflected boundaries, and all token gains. The extension bounds the effect of value variation using cumulative attention transport, identifying when the resulting mass increase compensates for this variation while attention on the original token set decreases. It therefore extends the analysis beyond components whose entire kernel neighborhood lies within content.

Increasing the smoothing width can further reduce the fraction returned to the original set, but it also increases transport distance and the associated coherence requirement. Gaussian width therefore controls a concrete balance between dilution and retention of shared value information. The proposition characterizes this balance for one attention operation under the stated coherence condition; its consequences for image generation are assessed empirically. Section~\ref{sec:reinforcement} tests content reinforcement within our method, and Sections~\ref{sec:components} and~\ref{sec:controls} examine the roles of allocation and guidance magnitude.

\subsection{Transfer to DeepFloyd IF}
IF uses T5 conditioning and attention over added text keys and image keys. We apply the same operator to Stage I's text-key block. Because T5 has no CLIP start token, smoothing spans the entire text block; content positions precede EOS, and positions after EOS receive $\lambda_p$. Image keys are unchanged, and Stage II uses its original attention. The transfer uses the SD-selected configuration without IF-specific strength tuning. This adaptation tests the design with a different text encoder and pixel-space denoising while retaining the same redistribution-and-reinforcement rule.

\section{Experiments}
\label{sec:experiments}
We evaluate whether redistribution and reinforcement can jointly reduce training-image similarity while retaining prompt-consistent generation. After benchmarking the suppression--quality trade-off, we test whether stronger content conditioning remains useful during mitigation (Figure~\ref{fig:reinforcement}), whether redistribution improves the trade-off at fixed token gains (Table~\ref{tab:ablation}), and whether the effect is explained by guidance magnitude alone (Table~\ref{tab:mechanism}). We then examine the EOS suppression boundary, transfer to a different model family, and selective application with an existing detector.

\subsection{Evaluation setup}
\begin{figure}[!t]
\centering
\includegraphics[width=\linewidth]{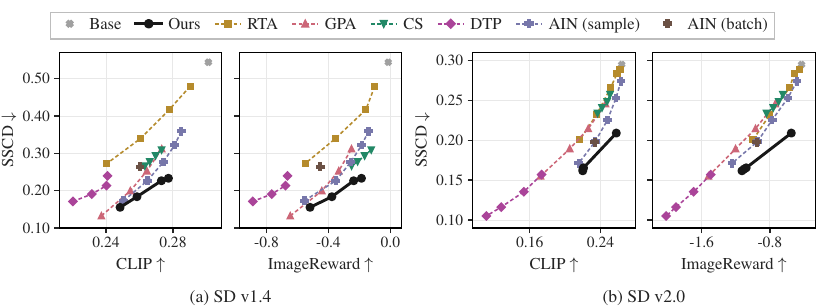}
\par\vspace{-6pt}
\caption{\textbf{Mitigation--quality trade-offs} on (a) SD v1.4 and (b) SD v2.0. All recorded operating points are shown; lines connect settings within each method. Lower-right is preferable.}
\label{fig:legacy}
\vspace{-6pt}
\end{figure}

\paragraph{Models and comparisons.}
The baseline benchmark uses 500 memorized prompts for SD v1.4 and 219 for SD v2.0 from prior work \citep{Memorization_500prompt,Memorization_unveiling}. We generate ten images per prompt with 50 steps and CFG scale 7, using DDIM \citep{DDIM} for v1.4 and Euler for v2.0. Baselines are random token addition (RTA; \citealp{Memorization_understanding}), gradient-based prompt adjustment (GPA; \citealp{Memorization_detecting}), cross-attention scaling (CS; \citealp{Memorization_unveiling}), opposite guidance with a dynamic transition point (DTP; \citealp{Memorization_basin}), and initial-noise adjustment (AIN; \citealp{AIN}). Figure~\ref{fig:legacy} includes every recorded operating point.

\paragraph{Design and mechanism studies.}
Controlled studies use SD v1.4 with the same DDIM schedule and CFG scale, producing $512\times512$ outputs. The EOS-boundary sweep covers 500 prompts with four images each; redistribution and guidance controls use a common subset of the same benchmark. Compared settings share initial random draws.

\paragraph{Metrics.}
We assess training-image similarity with SSCD \citep{SSCD}, prompt alignment with CLIP cosine similarity, and learned preference with ImageReward \citep{CLIPscore,Imagereward}. In the controlled SD studies, each generated image is compared with its prompt's available ground-truth references using the maximum SSCD, and scores are averaged over samples.

\subsection{Memorization--quality trade-offs}
\label{sec:external}
\textbf{Empirical Pareto-frontier performance.} In Figure~\ref{fig:legacy}, each of our four operating points lies on the empirical \emph{Pareto frontier} in both the SSCD--CLIP and SSCD--ImageReward comparisons, on both models: no evaluated alternative improves one objective without worsening the other. Varying the width traverses this trade-off, with smaller widths favoring quality and larger widths favoring stronger suppression. For example, our method ($\sigma=4$) and per-sample AIN (target 0.9) both report SSCD 0.2265 on SD v1.4; ours attains higher CLIP (0.2731 versus 0.2647) and ImageReward ($-0.2374$ versus $-0.3522$). Against the attention-based CS baseline, our $\sigma=1$ setting attains lower SSCD and higher CLIP than every evaluated CS setting on both models. On SD v2.0, it also achieves higher ImageReward.

\begin{figure}[tbp]
\centering
\includegraphics[width=4.6in]{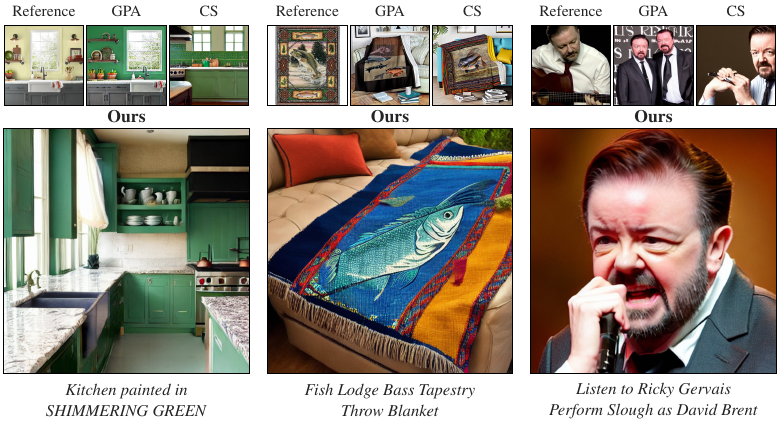}
\par\vspace{-6pt}
\caption{\textbf{Prompt-consistent alternatives.} Comparison with GPA and CS for three prompts, with training references shown for context. Our method retains the green kitchen, fish-patterned blanket, and performance context in alternative compositions.}
\label{fig:qualitative_semantics}
\vspace{-6pt}
\end{figure}

\paragraph{Prompt-consistent alternatives.}
Figure~\ref{fig:qualitative_semantics} compares how mitigation methods retain prompt content. For the performance prompt, GPA shows two people posing together, while CS shows a portrait of the subject. Our output depicts the subject performing with a handheld microphone, making the requested activity more explicit than in these baseline outputs while departing from the training reference's guitar-playing composition. The kitchen and blanket examples further illustrate alternative depictions that retain the requested green interior and fish motif.

\subsection{Reinforcing redistributed content}
\label{sec:reinforcement}
\begin{figure}[tbp]
\centering
\includegraphics[width=\linewidth]{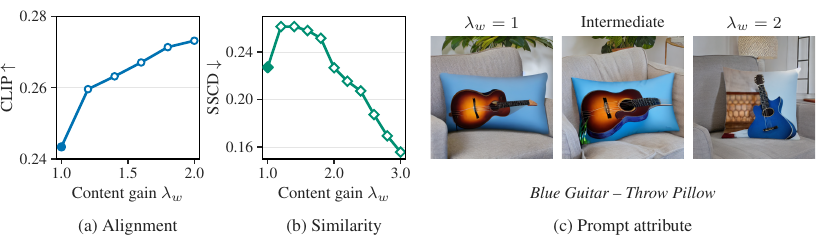}
\par\vspace{-6pt}
\caption{\textbf{Reinforcement supports prompt conditioning.} Content gain varies with redistribution and suppression retained: (a) prompt alignment (CLIP); (b) training-image similarity (SSCD). Filled markers indicate $\lambda_w=1$. (c) The guitar itself is blue at higher gain, matching the requested attribute.}
\label{fig:reinforcement}
\vspace{-6pt}
\end{figure}

With Gaussian redistribution and padding suppression retained, $\lambda_w=2$ yields higher CLIP than $\lambda_w=1$ at comparable SSCD, as shown in Figure~\ref{fig:reinforcement}. This improvement shows that stronger content conditioning can support prompt alignment without undoing the achieved suppression. Between these endpoints, SSCD responds non-monotonically: it first rises and then falls back to a comparable level at gain 2, while the wider sweep to 3 extends this decline. In the accompanying example, at $\lambda_w=1$ the pillow is blue but the guitar is brown; at $\lambda_w=2$, the guitar itself is blue, matching the requested attribute. These observations support reinforcement as a means of expressing useful prompt information during mitigation, without uniformly strengthening reproduction of the training reference. This generation-level behavior complements the local compatibility result in Eq.~\eqref{eq:gain_window}, which identifies when reinforcement preserves shared value information while redistribution reduces localized attention mass.

\subsection{How redistribution shapes the trade-off}
\label{sec:components}
To isolate the contribution of redistribution, we compare the main Gaussian operator with reweighting alone at fixed token gains. As a simple alternative, the uniform control averages each content token's attention with the within-span mean ($\rho=0.5$), preserving total content mass before weighting. We optionally normalize the weighted rows to examine the role of attention mass.

\begin{table}[!t]
\small
\centering
\setlength{\abovecaptionskip}{4pt}
\setlength{\belowcaptionskip}{4pt}
\caption{\textbf{Redistribution changes the trade-off under fixed token gains.} All intervention rows use $\lambda_w=2$, $\lambda_p=0.1$, and full intervention strength ($\alpha=1$). ``None'' removes redistribution only. ``Gaussian'' uses our redistribution operator; ``Uniform'' acts only within content. Our main method uses no row normalization.}
\label{tab:ablation}
\begin{tabular*}{0.88\linewidth}{@{\extracolsep{\fill}}llrrr@{}}
\toprule
Redistribution & Row norm. & SSCD $\downarrow$ & CLIP $\uparrow$ & ImageReward $\uparrow$\\
\midrule
None & No & 0.2052 & 0.2604 & $-0.3663$\\
\textbf{Gaussian (ours)}, $\sigma=4$ & No & 0.2088 & 0.2687 & $-0.2580$\\
Uniform (content) & No & 0.2279 & 0.2751 & $-0.2086$\\
\midrule
None & Yes & 0.3831 & 0.2913 & 0.0467\\
\textbf{Gaussian (ours)}, $\sigma=4$ & Yes & 0.3312 & 0.2793 & $-0.0231$\\
Uniform (content) & Yes & 0.3556 & 0.2875 & 0.0449\\
\midrule
Unmodified base & -- & 0.5465 & 0.3028 & 0.0594\\
\bottomrule
\end{tabular*}
\par\vspace{-3pt}
\end{table}

Without row normalization, Table~\ref{tab:ablation} shows that Gaussian redistribution improves both prompt alignment and image preference over reweighting alone at fixed token gains. This recovery accompanies only a small SSCD increase, from 0.2052 to 0.2088. This shows why redistribution is paired with reinforcement and suppression: it can recover generation quality while retaining most of the reduction in training-image similarity. Uniform redistribution also improves quality, but yields higher SSCD than Gaussian redistribution.

Row normalization moves all three settings toward higher alignment and preference but weaker suppression. With normalized rows, Gaussian redistribution instead reduces SSCD relative to reweighting alone, from 0.3831 to 0.3312, at a cost in alignment and preference. These comparisons show that redistribution shapes the trade-off under either normalization choice. Row normalization fixes attention mass but does not fix the magnitude of denoising guidance, motivating the following control.

\subsection{Guidance changes beyond magnitude attenuation}
\label{sec:controls}
To examine whether reduced guidance magnitude alone explains suppression, we compare same-state direction--norm controls. Write the CFG update as $\epsilon_u+w g$, where $g=\epsilon_c-\epsilon_u$ \citep{CFG}. At each control trajectory's current latent $x_k$, we recompute base and intervened guidance, $g_0(x_k)$ and $g_F(x_k)$, and form
\begin{equation}
 g_{0\to F}=\|g_F\|_2\frac{g_0}{\|g_0\|_2},\qquad
 g_{F\to0}=\|g_0\|_2\frac{g_F}{\|g_F\|_2}.
 \label{eq:swap}
\end{equation}
The two controls combine base direction with intervention norm, and intervention direction with base norm, respectively.

\begin{table}[!t]
\small
\centering
\setlength{\abovecaptionskip}{4pt}
\setlength{\belowcaptionskip}{4pt}
\caption{\textbf{Suppression persists after restoring guidance magnitude.} Direction and norm are controlled at the same latent state.}
\label{tab:mechanism}
\begin{tabular*}{0.88\linewidth}{@{\extracolsep{\fill}}llrrrr@{}}
\toprule
Direction & Norm & SSCD $\downarrow$ & CLIP $\uparrow$ & ImageReward $\uparrow$ & $\Delta$SSCD vs. base\\
\midrule
Base & Base & 0.5465 & 0.3028 & 0.0594 & --\\
Base & Intervention & 0.4988 & 0.2964 & 0.0150 & $-0.0477$\\
Intervention & Base & 0.3433 & 0.2826 & $-0.0179$ & $-0.2032$\\
Intervention & Intervention & 0.3458 & 0.2851 & 0.0245 & $-0.2007$\\
\bottomrule
\end{tabular*}
\par\vspace{-8pt}
\end{table}

In Table~\ref{tab:mechanism}, restoring the base guidance norm retains an SSCD reduction of 0.2032, close to 0.2007 when intervention direction and norm are used together; changing only the norm yields 0.0477. These controls show that scalar attenuation alone does not explain suppression: the combined attention intervention changes denoising direction in a way matched scalar attenuation does not reproduce. For the intervention direction, retaining its own norm gives higher CLIP and ImageReward than restoring the base norm at similar SSCD, showing that norm choice still matters for the quality trade-off.

Reinforcement improves alignment within the redistributed pattern, while redistribution recovers quality at fixed gains in the main unnormalized setting. Suppression beyond matched scalar attenuation further supports jointly controlling conditioning allocation and strength.

\subsection{Preserving the first EOS token}
\label{sec:eos}
\begin{figure}[tbp]
\centering
\setlength{\abovecaptionskip}{2pt}
\setlength{\belowcaptionskip}{0pt}
\includegraphics[width=4.8in]{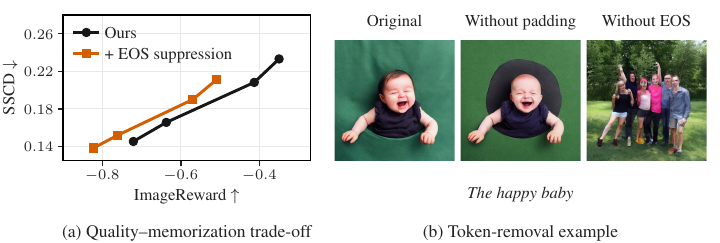}
\par
\caption{\textbf{Preserving EOS balances suppression with context.} (a) Our method with and without EOS attenuation. (b) In a separate token-removal example, removing EOS produces a group of people instead of the requested baby, whereas removing padding retains the baby depiction.}
\label{fig:eos}
\vspace{-4pt}
\end{figure}

Content reinforcement and selective suppression play complementary roles. Reducing the padding gain lowers both SSCD and CLIP (Appendix~\ref{app:reinforcement}), making the suppression boundary important for retaining useful conditioning. We therefore test whether attenuation should include the first EOS token. Figure~\ref{fig:eos}(a) varies its multiplier within our method. Preserving EOS yields favorable ImageReward--SSCD operating points across the measured sweep; extending attenuation to EOS reaches a lower similarity endpoint at a greater preference cost. At each fixed width, the additional suppression lowers both SSCD and ImageReward. The separate token-removal example in Figure~\ref{fig:eos}(b) illustrates the contextual information affected by this choice. Together, the quantitative trade-off and the qualitative change motivate treating the first EOS separately from the padding span.

\begin{table}[!t]
\small
\centering
\setlength{\abovecaptionskip}{4pt}
\setlength{\belowcaptionskip}{4pt}
\caption{\textbf{Suppression transfers to DeepFloyd IF.} Our SD-selected configuration is transferred without IF-specific tuning.}
\label{tab:if}
\begin{tabular}{lrrrr}
\toprule
IF setting & Template matches / 144 & Template SSCD $\downarrow$ & CLIP $\uparrow$ & ImageReward $\uparrow$\\
\midrule
Base & 56 (38.89\%) & 0.1985 & 0.3045 & 0.3988\\
\textbf{Ours} & 29 (20.14\%) & 0.1528 & 0.2842 & 0.1058\\
\bottomrule
\end{tabular}
\par\vspace{-8pt}
\end{table}

\subsection{Transfer to DeepFloyd IF}
\label{sec:if}
We test whether our SD-selected configuration transfers to T5-conditioned, pixel-space IF without target-specific tuning. Using template annotations \citep{Memorization_500prompt}, we evaluate 48 template-verbatim (TV) prompts across 11 shared-reference groups and three seeds (144 images per arm). As shown in Table~\ref{tab:if}, our method reduces template matches from 38.89\% to 20.14\%. This reduction under the SD-selected settings shows that the intervention can suppress reproduction with a different text encoder and generation space. CLIP and ImageReward also decrease relative to the base model, so the transferred configuration retains a suppression benefit with a quality cost.

\subsection{Selective mitigation with an existing detector}
To limit intervention on ordinary prompts, we intervene at full strength ($\sigma=4$) only on inputs flagged by the initial-step guidance detector of \citet{Memorization_detecting}. At a calibrated 5\% false-positive rate, mean SSCD on 500 memorized prompts falls from 0.5486 to 0.3475, versus 0.2078 when every prompt is treated. Average ImageReward changes by $-0.0065$ over normal prompts and $-0.1291$ within the flagged normal subset. Because unflagged inputs follow the original generation path, selective application preserves most normal generations while still reducing reproduction. Its weaker suppression than universal application reflects the memorization-prone prompts left untreated.

\section{Conclusion}
We show that stronger content conditioning within redistributed attention can improve prompt alignment at comparable training-image similarity, making reinforcement an active means of generating useful alternatives during mitigation. The local analysis identifies conditions under which reinforcement preserves shared value information while redistribution dilutes localized attention, and controlled comparisons show quality recovery from redistribution at fixed token gains. Our method attains empirical Pareto-frontier performance on two Stable Diffusion benchmarks. An SD-selected configuration also reduces template reproduction in DeepFloyd IF without further tuning. The resulting principle is to reinforce useful prompt content during mitigation, using redistribution to improve the balance between reproduction suppression and prompt-consistent generation.

\clearpage
\subsection*{AI use statement}
We primarily used generative AI tools for manuscript editing and figure and table formatting. We also used these tools for reviewing mathematical claims and proofs for completeness and consistency, including feedback on assumptions and proof arguments. These tools also assisted with writing experimental code and provided feedback on ablation studies. The authors directed the revisions and take responsibility for the final content, claims, and results of this work.

\bibliography{foca_references}

@String(CVPR= {IEEE Conf. Comput. Vis. Pattern Recog.})

@String(TOG= {ACM Trans. Graph.})

@String(CVPR  = {CVPR})

@String(TOG   = {ACM TOG})

@inproceedings{SD,
  title={High-resolution image synthesis with latent diffusion models},
  author={Rombach, Robin and Blattmann, Andreas and Lorenz, Dominik and Esser, Patrick and Ommer, Bj{\"o}rn},
  booktitle={Proceedings of the IEEE/CVF conference on computer vision and pattern recognition},
  pages={10684--10695},
  year={2022}
}

@article{Dall-E,
  title={Hierarchical Text-Conditional Image Generation with {CLIP} Latents},
  author={Ramesh, Aditya and Dhariwal, Prafulla and Nichol, Alex and Chu, Casey and Chen, Mark},
  journal={arXiv preprint arXiv:2204.06125},
  year={2022}
}

@inproceedings{Memorization_usenix,
  title={Extracting training data from diffusion models},
  author={Carlini, Nicholas and Hayes, Jamie and Nasr, Milad and Jagielski, Matthew and Sehwag, Vikash and Tram{\`e}r, Florian and Balle, Borja and Ippolito, Daphne and Wallace, Eric},
  booktitle={32nd USENIX Security Symposium (USENIX Security 23)},
  pages={5253--5270},
  year={2023}
}

@article{LAION,
  title={{LAION-5B}: An open large-scale dataset for training next generation image-text models},
  author={Schuhmann, Christoph and Beaumont, Romain and Vencu, Richard and Gordon, Cade and Wightman, Ross and Cherti, Mehdi and Coombes, Theo and Katta, Aarush and Mullis, Clayton and Wortsman, Mitchell and others},
  journal={Advances in neural information processing systems},
  volume={35},
  pages={25278--25294},
  year={2022}
}

@inproceedings{Memorization_forgery,
  title={Diffusion art or digital forgery? investigating data replication in diffusion models},
  author={Somepalli, Gowthami and Singla, Vasu and Goldblum, Micah and Geiping, Jonas and Goldstein, Tom},
  booktitle={Proceedings of the IEEE/CVF conference on computer vision and pattern recognition},
  pages={6048--6058},
  year={2023}
}

@inproceedings{Memorization_unveiling,
  title={Unveiling and mitigating memorization in text-to-image diffusion models through cross attention},
  author={Ren, Jie and Li, Yaxin and Zeng, Shenglai and Xu, Han and Lyu, Lingjuan and Xing, Yue and Tang, Jiliang},
  booktitle={European Conference on Computer Vision},
  pages={340--356},
  year={2024},
  organization={Springer}
}

@article{Memorization_understanding,
  title={Understanding and mitigating copying in diffusion models},
  author={Somepalli, Gowthami and Singla, Vasu and Goldblum, Micah and Geiping, Jonas and Goldstein, Tom},
  journal={Advances in Neural Information Processing Systems},
  volume={36},
  pages={47783--47803},
  year={2023}
}

@inproceedings{Memorization_detecting,
  title={Detecting, explaining, and mitigating memorization in diffusion models},
  author={Wen, Yuxin and Liu, Yuchen and Chen, Chen and Lyu, Lingjuan},
  booktitle={The Twelfth International Conference on Learning Representations},
  year={2024}
}

@article{Memorization_nemo,
  title={Finding {NeMo}: Localizing neurons responsible for memorization in diffusion models},
  author={Hintersdorf, Dominik and Struppek, Lukas and Kersting, Kristian and Dziedzic, Adam and Boenisch, Franziska},
  journal={Advances in Neural Information Processing Systems},
  volume={37},
  pages={88236--88278},
  year={2024}
}

@InProceedings{Memorization_basin,
    author    = {Jain, Anubhav and Kobayashi, Yuya and Shibuya, Takashi and Takida, Yuhta and Memon, Nasir and Togelius, Julian and Mitsufuji, Yuki},
    title     = {Classifier-Free Guidance Inside the Attraction Basin May Cause Memorization},
    booktitle = {Proceedings of the IEEE/CVF Conference on Computer Vision and Pattern Recognition (CVPR)},
    month     = {June},
    year      = {2025},
    pages     = {12871-12879}
}

@article{Memorization_500prompt,
  title={A reproducible extraction of training images from diffusion models},
  author={Webster, Ryan},
  journal={arXiv preprint arXiv:2305.08694},
  year={2023}
}

@article{Cross_excite,
  title={{Attend-and-Excite}: Attention-based semantic guidance for text-to-image diffusion models},
  author={Chefer, Hila and Alaluf, Yuval and Vinker, Yael and Wolf, Lior and Cohen-Or, Daniel},
  journal={ACM Transactions on Graphics (TOG)},
  volume={42},
  number={4},
  pages={1--10},
  year={2023},
  publisher={ACM New York, NY, USA}
}

@article{Cross_syngen,
  title={Linguistic binding in diffusion models: Enhancing attribute correspondence through attention map alignment},
  author={Rassin, Royi and Hirsch, Eran and Glickman, Daniel and Ravfogel, Shauli and Goldberg, Yoav and Chechik, Gal},
  journal={Advances in Neural Information Processing Systems},
  volume={36},
  pages={3536--3559},
  year={2023}
}

@inproceedings{Cross_structure_diffusion,
title={Training-Free Structured Diffusion Guidance for Compositional Text-to-Image Synthesis},
author={Weixi Feng and Xuehai He and Tsu-Jui Fu and Varun Jampani and Arjun Reddy Akula and Pradyumna Narayana and Sugato Basu and Xin Eric Wang and William Yang Wang},
booktitle={The Eleventh International Conference on Learning Representations },
year={2023},
url={https://openreview.net/forum?id=PUIqjT4rzq7}
}

@inproceedings{Cross_prompttoprompt,
title={Prompt-to-Prompt Image Editing with Cross-Attention Control},
author={Amir Hertz and Ron Mokady and Jay Tenenbaum and Kfir Aberman and Yael Pritch and Daniel Cohen-Or},
booktitle={The Eleventh International Conference on Learning Representations },
year={2023},
url={https://openreview.net/forum?id=_CDixzkzeyb}
}

@article{DDPM,
  title={Denoising diffusion probabilistic models},
  author={Ho, Jonathan and Jain, Ajay and Abbeel, Pieter},
  journal={Advances in neural information processing systems},
  volume={33},
  pages={6840--6851},
  year={2020}
}

@inproceedings{DDIM,
title={Denoising Diffusion Implicit Models},
author={Jiaming Song and Chenlin Meng and Stefano Ermon},
booktitle={International Conference on Learning Representations},
year={2021},
url={https://openreview.net/forum?id=St1giarCHLP}
}

@article{Imagen,
  title={Photorealistic text-to-image diffusion models with deep language understanding},
  author={Saharia, Chitwan and Chan, William and Saxena, Saurabh and Li, Lala and Whang, Jay and Denton, Emily L and Ghasemipour, Kamyar and Gontijo Lopes, Raphael and Karagol Ayan, Burcu and Salimans, Tim and others},
  journal={Advances in neural information processing systems},
  volume={35},
  pages={36479--36494},
  year={2022}
}

@inproceedings{EOS_good,
title={Towards Understanding the Working Mechanism of Text-to-Image Diffusion Model},
author={Mingyang Yi and Aoxue Li and Yi Xin and Zhenguo Li},
booktitle={The Thirty-eighth Annual Conference on Neural Information Processing Systems},
year={2024},
url={https://openreview.net/forum?id=zTu0QEpvtZ}
}

@inproceedings{Memorization_LID,
title={A Geometric Framework for Understanding Memorization in Generative Models},
author={Brendan Leigh Ross and Hamidreza Kamkari and Tongzi Wu and Rasa Hosseinzadeh and Zhaoyan Liu and George Stein and Jesse C. Cresswell and Gabriel Loaiza-Ganem},
booktitle={The Thirteenth International Conference on Learning Representations},
year={2025},
url={https://openreview.net/forum?id=aZ1gNJu8wO}
}

@inproceedings{CLIPscore,
  title={Learning transferable visual models from natural language supervision},
  author={Radford, Alec and Kim, Jong Wook and Hallacy, Chris and Ramesh, Aditya and Goh, Gabriel and Agarwal, Sandhini and Sastry, Girish and Askell, Amanda and Mishkin, Pamela and Clark, Jack and others},
  booktitle={International conference on machine learning},
  pages={8748--8763},
  year={2021},
  organization={PMLR}
}

@inproceedings{SSCD,
  title={A self-supervised descriptor for image copy detection},
  author={Pizzi, Ed and Roy, Sreya Dutta and Ravindra, Sugosh Nagavara and Goyal, Priya and Douze, Matthijs},
  booktitle={Proceedings of the IEEE/CVF Conference on Computer Vision and Pattern Recognition},
  pages={14532--14542},
  year={2022}
}

@inproceedings{CFG,
title={Classifier-Free Diffusion Guidance},
author={Jonathan Ho and Tim Salimans},
booktitle={NeurIPS 2021 Workshop on Deep Generative Models and Downstream Applications},
year={2021},
url={https://openreview.net/forum?id=qw8AKxfYbI}
}

@inproceedings{SDE1,
title={Score-Based Generative Modeling through Stochastic Differential Equations},
author={Yang Song and Jascha Sohl-Dickstein and Diederik P Kingma and Abhishek Kumar and Stefano Ermon and Ben Poole},
booktitle={International Conference on Learning Representations},
year={2021},
url={https://openreview.net/forum?id=PxTIG12RRHS}
}

@inproceedings{Appendix_clip,
  title={Reproducible scaling laws for contrastive language-image learning},
  author={Cherti, Mehdi and Beaumont, Romain and Wightman, Ross and Wortsman, Mitchell and Ilharco, Gabriel and Gordon, Cade and Schuhmann, Christoph and Schmidt, Ludwig and Jitsev, Jenia},
  booktitle={Proceedings of the IEEE/CVF conference on computer vision and pattern recognition},
  pages={2818--2829},
  year={2023}
}

@article{DISC2021,
  title={The 2021 image similarity dataset and challenge},
  author={Douze, Matthijs and Tolias, Giorgos and Pizzi, Ed and Papakipos, Zo{\"e} and Chanussot, Lowik and Radenovic, Filip and Jenicek, Tomas and Maximov, Maxim and Leal-Taix{\'e}, Laura and Elezi, Ismail and others},
  journal={arXiv preprint arXiv:2106.09672},
  year={2021}
}

@article{Imagereward,
  title={{ImageReward}: Learning and evaluating human preferences for text-to-image generation},
  author={Xu, Jiazheng and Liu, Xiao and Wu, Yuchen and Tong, Yuxuan and Li, Qinkai and Ding, Ming and Tang, Jie and Dong, Yuxiao},
  journal={Advances in Neural Information Processing Systems},
  volume={36},
  pages={15903--15935},
  year={2023}
}

@inproceedings{
    AIN,
    title={Adjusting Initial Noise to Mitigate Memorization in Text-to-Image Diffusion Models},
    author={Hyeonggeun Han and Sehwan Kim and Hyungjun Joo and Sangwoo Hong and Jungwoo Lee},
    booktitle={The Thirty-ninth Annual Conference on Neural Information Processing Systems},
    year={2025},
    url={https://openreview.net/forum?id=4KengZ1RNX}
}

@inproceedings{chenchen,
  title={Towards memorization-free diffusion models},
  author={Chen, Chen and Liu, Daochang and Xu, Chang},
  booktitle={Proceedings of the IEEE/CVF Conference on Computer Vision and Pattern Recognition},
  pages={8425--8434},
  year={2024}
}

@inproceedings{BrightEnding,
  title={Exploring Local Memorization in Diffusion Models via Bright Ending Attention},
  author={Chen, Chen and Liu, Daochang and Shah, Mubarak and Xu, Chang},
  booktitle={The Thirteenth International Conference on Learning Representations},
  year={2025},
  url={https://arxiv.org/abs/2410.21665}
}
\bibliographystyle{iclr2027_conference}
\clearpage
\appendix
\section{Analysis of redistribution and reinforcement}
\label{app:theory}

\subsection{Operator and scope}
The analysis holds the input attention row and value vectors fixed at one cross-attention operation. It therefore analyzes the transformation applied by our method, before subsequent nonlinear denoising. An attention component supported on $\mathcal T$ can be isolated from any row by $a_i^{\mathcal T}=a_i\mathbf1\{i\in\mathcal T\}$. No exact zeros in a softmax row are required. Since the main operator is linear in the post-softmax weights, the complete transformed row is the sum of its transformed components.

For clarity, columns represent attention vectors in this appendix. Let $G$ be the matrix of the implemented blur: $b=Ga$. Thus $G_{ji}$ is the contribution of donor position $i$ to output position $j$. On positions $1,\ldots,L-1$, the code uses a normalized truncated Gaussian with reflection padding; the start token is unchanged. Its matrix satisfies $G\mathbf1=\mathbf1$, but its column sums can differ from one. Consequently, the full reflected operator need not preserve total attention mass. Inside the local region specified in Proposition~\ref{prop:reinforcement}, reflection is inactive and $G_{ji}=g_{j-i}$ on the component's support, with unit column sums there. The kernel neighborhood must not touch either reflected endpoint; mere inclusion of an endpoint is insufficient.

With row $j$ of $V$ equal to $v_j^\top$, the transformed values admit the exact identity
\begin{equation}
 \widehat o^\top=(1-\alpha)a^\top V+\alpha a^\top G^\top\Lambda V.
 \label{eq:effective_values}
\end{equation}
The identity expresses the method as a change in the mixtures of value vectors seen by the original attention weights. Token gains act on the recipients of redistribution. It is valid for the full reflected operator and explains why amplifying the redistributed pattern need not recover the original conditioning.

\subsection{Proof of Proposition~\ref{prop:reinforcement}}
Write $R$ for the kernel radius. For every donor $i\in\mathcal T$, all recipients $i-R,\ldots,i+R$ lie in the content span and avoid reflection, so $b_j=\sum_{i\in\mathcal T}g_{j-i}a_i$ and $\sum_j g_{j-i}=1$. Because all recipients have the same gain, the transformed component is exactly
\begin{equation}
 \widehat a=(1-\alpha)a+\alpha\lambda_w b.
\end{equation}
Using $s_j=u^\top v_j$ and the local coherence assumption,
\begin{align}
 |S(b)-S(a)|
 &=\left|\sum_{i\in\mathcal T}a_i\sum_jg_{j-i}(s_j-s_i)\right|\\
 &\leq\sum_{i\in\mathcal T}a_i\sum_jg_{j-i}\ell|j-i|
 =c\ell\mu_\sigma.
\end{align}
Since $S(a)\geq cs_*>0$, this gives $S(b)\geq(1-\varepsilon)S(a)$. Applying the blend yields Eq.~\eqref{eq:semantic_bound}. Independently,
\begin{equation}
 M_{\mathcal T}(b)=\sum_{i\in\mathcal T}a_i\sum_{j\in\mathcal T}g_{j-i}
 \leq\kappa_{\mathcal T}c,
\end{equation}
which gives Eq.~\eqref{eq:localized_bound}. The Gaussian's largest coefficient is $g_0$, so summing over $|\mathcal T|$ recipients gives $\kappa_{\mathcal T}\leq|\mathcal T|g_0$; normalization also gives $\kappa_{\mathcal T}\leq1$. Finally, $\varepsilon+\kappa_{\mathcal T}<1$ makes the interval in Eq.~\eqref{eq:gain_window} nonempty. Its lower endpoint ensures $\lambda_w(1-\varepsilon)\geq1$ and its strict upper endpoint ensures $\lambda_w\kappa_{\mathcal T}<1$. With $\alpha>0$, the two conclusions follow.\hfill$\square$

\paragraph{Localized value information.}
One precise interpretation of the mass bound is a value feature restricted to $\mathcal T$. Suppose, for a local response model, $v_j=\bar v_j+\mathbf1\{j\in\mathcal T\}d$ for a shared residual vector $d$. Its contribution from the component is initially $cd$, and after intervention is $M_{\mathcal T}(\widehat a)d$. Equation~\eqref{eq:localized_bound} bounds its attenuation, while Eq.~\eqref{eq:semantic_bound} preserves a locally coherent projection. This residual provides a structural model of token-localized information. Generation-level effects are evaluated separately using training-image similarity and quality.

\subsection{A transport bound for the complete attention row}
\label{app:full_row_theory}
We now retain the complete attention row, including background attention, reflection, and all token gains. This removes the requirement that the kernel neighborhood lie inside content. Let $p=a/(\mathbf1^\top a)$, $z=Gp$, $b_\lambda=\Lambda_\lambda z$, and $r_\lambda=\mathbf1^\top b_\lambda>0$, where $\lambda=\lambda_w$ and the other gains remain fixed. Normalizing $a$ here only defines analysis variables; the intervention retains its changed row mass. Define the cumulative mismatch
\begin{equation}
 D_k(\lambda)=r_\lambda\sum_{i=0}^{k}p_i-\sum_{i=0}^{k}b_{\lambda,i},
 \qquad k=0,\ldots,L-2.
 \label{eq:cumulative_mismatch}
\end{equation}
For nonnegative adjacent-variation budgets $\eta_k$, consider normalized projected features in
\begin{equation}
 \mathcal F_{\boldsymbol\eta}(p)=
 \{f\in\mathbb R^L:p^\top f=1,\quad |f_{k+1}-f_k|\leq\eta_k\ \text{for every }k\}.
 \label{eq:full_row_coherence}
\end{equation}
These budgets allow different degrees of coherence within content and across special-token boundaries.

\begin{proposition}[Complete-row preservation and localized dilution]
\label{prop:full_row_transport}
For the complete reflected operator, the sharp lower bound over this feature class is
\begin{equation}
 \inf_{f\in\mathcal F_{\boldsymbol\eta}(p)}b_\lambda^\top f
 =r_\lambda-\sum_{k=0}^{L-2}\eta_k|D_k(\lambda)|.
 \label{eq:full_row_transport_bound}
\end{equation}
Let $s_i=u^\top v_i$ for a unit direction $u$, $\mu=p^\top s>0$, and $s/\mu\in\mathcal F_{\boldsymbol\eta}(p)$. For $\mathcal T\subset\mathcal C$, write $R_0=\sum_{i\in\mathcal T}p_i>0$ and $R_B=\sum_{j\in\mathcal T}z_j$. Every $0<\alpha\leq1$ preserves $S(a)$ and strictly reduces $M_{\mathcal T}(a)$ if
\begin{equation}
 r_\lambda-\sum_k\eta_k|D_k(\lambda)|\geq1,
 \qquad \lambda R_B<R_0.
 \label{eq:full_row_certificate}
\end{equation}
The first condition is necessary and sufficient for preservation uniformly over the specified feature class; the second is the exact condition for reducing attention on $\mathcal T$.
\end{proposition}

\paragraph{Proof.}
The vector $r_\lambda p-b_\lambda$ sums to zero. Summation by parts gives
\begin{equation}
 b_\lambda^\top f-r_\lambda p^\top f
 =\sum_kD_k(\lambda)(f_{k+1}-f_k)
 \geq-\sum_k\eta_k|D_k(\lambda)|.
\end{equation}
Choose $f_{k+1}-f_k=-\eta_k\operatorname{sign}(D_k)$ and add a constant to achieve $p^\top f=1$; this attains equality. Blending then gives
\begin{align}
 \frac{S(\widehat a)}{S(a)}
 &\geq1-\alpha+\alpha\left[r_\lambda-\sum_k\eta_k|D_k(\lambda)|\right],\\
 \frac{M_{\mathcal T}(\widehat a)}{M_{\mathcal T}(a)}
 &=1-\alpha+\alpha\lambda R_B/R_0.
\end{align}
The second identity includes all donor contributions through $z=Gp$. These relations prove the stated conditions.\hfill$\square$

The first inequality in Eq.~\eqref{eq:full_row_certificate} requires the mass surplus $r_\lambda-1$ to cover the feature variation induced by redistribution. With a common budget $\eta_k=\eta$, the cost is $\eta\tau_\lambda$, where $\tau_\lambda=\sum_k|D_k(\lambda)|$ measures cumulative transport on token positions. The second inequality independently limits retention on the original set. For a fixed row and blur, $b_\lambda=\lambda x+y$ for fixed $x,y$, so $r_\lambda$ and $D_k(\lambda)$ are affine in $\lambda$. The preservation bound is therefore concave and piecewise affine. Its feasible gains, intersected with $\lambda\geq1$ and the dilution condition, form an interval that may be empty.

\paragraph{A short-content example.}
Take $L=77$, $\mathcal C=\{1,2,3,4\}$, first EOS at position 5, $\mathcal T=\{2\}$, and $p=0.9e_2+(0.1/77)\mathbf1$, where $e_2$ is the unit vector at position 2. Use the implemented reflected Gaussian with $\sigma=4$ and radius 12, $\lambda_w=2$, $\lambda_p=0.1$, and unit start/EOS gains. Direct calculation gives $r_\lambda=1.42078$, $\tau_\lambda=5.55221$, and $\lambda R_B/R_0=0.37849$. At full mix, every feature with $\eta_k=0.07$ has response ratio at least $1.42078-0.07(5.55221)=1.0321$, while the original token retains only $0.37849$ of its attention. The local proposition cannot cover this four-token span because its kernel extends beyond content. This constructed example demonstrates a nonempty preservation--dilution regime under the complete operator.

\paragraph{Exact diagnostic for fixed values.}
When the actual projected values are available, define $S_C=\sum_{j\in\mathcal C}z_js_j$ and $S_R=\sum_{j\notin\mathcal C}\lambda_jz_js_j$, where $\lambda_j=1$ for start/EOS and $\lambda_p$ for padding. Then $b_\lambda^\top s=\lambda S_C+S_R$. For $S_C>0$ and $R_B>0$, the exact gain interval preserving this particular projection and diluting $\mathcal T$ is
\begin{equation}
 \max\!\left\{1,\frac{\mu-S_R}{S_C}\right\}
 \leq\lambda<\frac{R_0}{R_B}.
 \label{eq:general_gain_window}
\end{equation}
If $S_C<0$ the preservation inequality reverses, while $S_C=0$ makes it independent of gain; $R_B=0$ removes the finite dilution upper bound. Equation~\eqref{eq:full_row_certificate} instead supplies a structural guarantee from bounded feature variation, without fixing a particular profile. These expressions characterize admissible gains for a fixed attention operation. Our main method uses the common content gain $\lambda_w=2$.

\subsection{Implications and numerical verification}
\paragraph{Complementary roles of redistribution and gain.}
For the same local component, setting $\alpha=1$ in Eqs.~\eqref{eq:semantic_bound}--\eqref{eq:localized_bound} gives, when $\varepsilon+\kappa_{\mathcal T}<1$,
\begin{equation}
 \frac{M_{\mathcal T}(\widehat a)/c}{S(\widehat a)/S(a)}
 \leq\frac{\kappa_{\mathcal T}}{1-\varepsilon}<1.
 \label{eq:relative_retention}
\end{equation}
The numerator measures retention on the original token set, and the denominator measures retention of the projected feature. The common gain cancels in this ratio: redistribution changes their relative retention, while reinforcement scales both responses. Choosing the gain within Eq.~\eqref{eq:gain_window} turns this relative advantage into absolute feature preservation and localized dilution. Amplification alone has $\kappa_{\mathcal T}=1$ and offers no such interval.

\paragraph{Kernel calculations.}
For an interior singleton, $\kappa_{\mathcal T}=g_0$. The implemented kernels at $\sigma=1$ and $\sigma=4$ have $(g_0,\mu_\sigma)=(0.39905,0.72669)$ and $(0.09991,3.15669)$, respectively. At full mix and $\lambda_w=2$, only $0.79810$ or $0.19982$ of that component's original coefficient returns to its source token, even though its total content mass doubles. These calculations illustrate why gain and localized influence can move in opposite directions. For the fixed gain $\lambda_w=2$, sufficient local conditions are $\varepsilon\leq1/2$ and $\kappa_{\mathcal T}<1/2$. For a wider support, the relevant quantity is $\kappa_{\mathcal T}$, not the singleton coefficient.

\paragraph{Relation to the experimental design.}
The gain interval couples dilution to coherence. Wider smoothing can reduce retention on a localized set while increasing the transport-distance bound on feature distortion. Content reinforcement compensates for the latter within the admissible interval. After the early smoothing window, $G=I$ and the local interval disappears; subsequent reweighting acts on attention along the already modified trajectory. The analysis therefore motivates early redistribution coupled with reinforcement, while the reported schedule and gains are empirical design choices. The gain sweep and the row-mass and direction/norm controls assess the resulting allocation--strength trade-off through prompt alignment, image preference, and training-image similarity.

\paragraph{Computational checks.}
Numerical verification covers the finite-kernel inequalities, complete-row gain conditions, and transport bound, including equality-achieving profiles. Independent linear-programming checks confirm the sharp bound in Eq.~\eqref{eq:full_row_transport_bound}. Verification scripts and detailed numerical results are included with the manuscript source.

Four counterexamples delimit the analysis: useful information confined to the same isolated token can be attenuated; a uniform attention pattern can be amplified everywhere; reflection can change total row mass; and background inflow can outweigh a localized component's dilution. These cases motivate the coherence, concentration, and full-row conditions used above.

\FloatBarrier
\Needspace{8\baselineskip}
\section{Implementation and reproducibility details}
\label{app:implementation}
The baseline benchmark was run on an A100 with PyTorch 1.13.0, torchvision 0.14.0, CUDA 11.7, and diffusers 0.26.3. The controlled SD and IF studies used an H100 80GB with PyTorch 2.5.1+cu121, diffusers 0.40.0, and transformers 5.12.1. ImageReward ran in a separate environment with transformers 4.30.2. The scorer for these studies uses OpenCLIP ViT-B-32 \citep{Appendix_clip} with the \texttt{laion2b\_s34b\_b79k} checkpoint \citep{LAION}; SSCD uses \texttt{sscd\_disc\_mixup} \citep{SSCD,DISC2021}; preference evaluation uses ImageReward v1.0. Scores are computed against the original prompt. Comparisons are made within each environment and sampling protocol. We report paired percentile bootstrap CIs without multiplicity adjustment. The baseline benchmark and EOS sweep provide aggregate comparisons; prompt-level CIs are reported for the controlled SD studies.

The controlled SD studies use DDIM \citep{DDIM} with 50 steps, CFG scale 7, and $512\times512$ outputs. Redistribution and guidance controls use a common 340-prompt subset of the 500-prompt benchmark, comprising 264 template-memorization (TM) and 76 verbatim-memorization (VM) prompts. The subset is stratified by memorization type, with identical captions and shared recorded ground-truth identifiers kept in the same partition. These controls use two seeds and four images per prompt and seed, with matched initial random draws across settings. Paired SD intervals first average samples and seeds within each prompt, then resample prompts with replacement 10,000 times. A high-similarity batch is defined by at least one of its four images exceeding SSCD 0.5.

A reproduction check compares the baseline benchmark with the H100 environment. The benchmark SD v1.4 base reports SSCD/CLIP/ImageReward 0.5439/0.3012/$-0.0140$, versus 0.5461/0.3021/$-0.0164$ in the H100 reproduction. For our method at $\sigma=4$, the corresponding triplets are 0.2265/0.2731/$-0.2374$ and 0.2348/0.2706/$-0.2794$. The base results agree closely; the larger difference in mitigated ImageReward motivates reporting within-protocol comparisons separately.

\begin{figure}[!htbp]
\centering
\includegraphics[width=\linewidth]{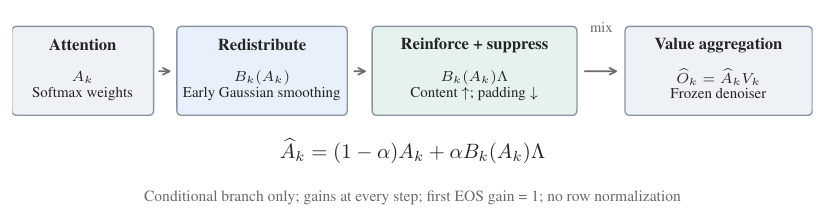}
\par\vspace{-3pt}
\caption{\textbf{Reinforcing redistributed attention.} Gaussian smoothing precedes content reinforcement and padding attenuation. The original and modified attention are mixed with strength $\alpha$ before value aggregation.}
\label{fig:overview}
\vspace{-3pt}
\end{figure}

Our method uses a reflected convolution over positions $1,\ldots,L-1$ in SD. Its kernel size is $2\lfloor3\sigma+0.5\rfloor+1$, reduced to the largest admissible odd size when necessary. The positional content kernel instead normalizes each donor row of a finite matrix over the content span. These operators have different boundary behavior and token support even at the same nominal width. All controlled SD settings operate on post-softmax probabilities; the moderate-strength comparison uses no temperature scaling. Figure~\ref{fig:overview} summarizes the operator, and Table~\ref{tab:implementation_mapping} maps paper labels to implementation settings.

The moderate-strength comparison covers 8,160 generated images across two settings; the IF evaluation covers 480 images across three settings, including the supplementary quality set.

\begin{table}[!htbp]
\small
\caption{Mapping between paper labels and implementation settings.}
\label{tab:implementation_mapping}
\begin{center}
\begin{tabular}{ll}
\toprule
Paper notation & Implementation identifier\\
\midrule
Ours ($\sigma=4$, $\alpha=1$) & \texttt{published\_s4} / full-strength Gaussian\\
\posvariant & \texttt{abl\_position\_norm1}\\
Ours ($\sigma=4$, $\alpha=0.625$) & \texttt{foca\_s4\_t0p625}\\
Content-uniform & \texttt{content\_uniform\_t1p5}\\
Base direction / positional norm & \texttt{control\_base\_direction}\\
Positional direction / base norm & \texttt{control\_foca\_direction}\\
\bottomrule
\end{tabular}
\end{center}

\vspace{-3pt}
\end{table}

\FloatBarrier
\section{Component variations and exact settings}
\label{app:variants}
For the ablations, row normalization is $\mathcal N_1(F)_{ij}=F_{ij}/\sum_\ell F_{i\ell}$, with $\mathcal N_0(F)=F$. We evaluate
$\widehat A_k=(1-\alpha)A_k+\alpha\mathcal N_\nu(B_k(A_k)\Lambda)$.
The main Gaussian method uses $\nu=0$. Table~\ref{tab:operators} lists the experimental variants used to study its components.

To separate redistribution within the content span from mixing across token classes, we also use
\begin{equation}
 B_k(A_k)_{i,\mathcal C}
 =(1-\rho)A_{k,i,\mathcal C}+\rho A_{k,i,\mathcal C}K,
 \label{eq:content}
\end{equation}
leaving other columns unchanged before reweighting. $K$ is a nonnegative row-stochastic donor-to-recipient matrix, so Eq.~\eqref{eq:content} preserves each row's total content mass. For positional redistribution,
$K_{uv}\propto\exp[-(u-v)^2/(2\sigma_c^2)]$, normalized over recipient $v$. For uniform redistribution, $K_{uv}=1/|\mathcal C|$. The latter moves content attention toward its within-span mean without adding uniform mass to special or padding tokens. It differs from uniform smoothing across the entire token sequence.

Normalization is a separate intervention axis. With reweighting alone,
\begin{equation}
 \mathcal N_1\!\left(\operatorname{softmax}(\ell)\Lambda\right)
 =\operatorname{softmax}(\ell+\log\lambda)
\end{equation}
for positive weights. Row normalization fixes attention mass while allowing the value mixture, denoiser output, and CFG magnitude to change.

\begin{table}[!htbp]
\small
\setlength{\tabcolsep}{4pt}
\caption{Settings for the main method and component variations. ``Global'' excludes SD's start token. In all controlled SD settings, redistribution acts in the first 5 of 50 steps and weighting acts in all 50. \posvariant{} is used for mechanism analysis; our moderate-strength setting and Content-uniform form the fixed-setting transfer comparison. The uniform rows in Table~\ref{tab:ablation} instead use its common gains $\lambda_w=2$ and $\lambda_p=0.1$.}
\label{tab:operators}
\begin{center}
\begin{tabular}{llccccc}
\toprule
Setting & Redistribution & $\rho$ & $\alpha$ & $\lambda_w$ & $\lambda_p$ & Norm.\\
\midrule
Ours (benchmark) & Global Gaussian, $\sigma\in\{1,4,10,100\}$ & -- & 1 & 2 & 0.1 & No\\
\posvariant & Content positional, $\sigma_c=1$ & 1 & 1 & 2 & 0.1 & Yes\\
Ours (moderate) & Global Gaussian, $\sigma=4$ & -- & 0.625 & 2 & 0.1 & No\\
Content-uniform & Content uniform & 0.5 & 1 & $2^{1.5}$ & $0.1^{1.5}$ & Yes\\
\bottomrule
\end{tabular}
\end{center}
\vspace{-3pt}
\end{table}

\FloatBarrier
\paragraph{Direct ablation of Gaussian redistribution.}
The unnormalized Gaussian comparison in Table~\ref{tab:ablation_full} uses the same 340 prompts, two seeds, and four images per prompt per seed as the content-only controls. Its three settings use $\lambda_w=2$, $\lambda_p=0.1$, full intervention strength ($\alpha=1$), and no row normalization, varying only Gaussian redistribution: disabled, $\sigma=1$, or $\sigma=4$. In this unnormalized comparison, the aggregate quality gains are mainly driven by TM prompts; SSCD increases on TM and decreases on VM. The narrower kernel favors greater quality recovery, whereas $\sigma=4$ retains stronger suppression.

The $\sigma=4$ No/Yes pair in Table~\ref{tab:ablation} holds the Gaussian width and token gains fixed to isolate the effect of normalizing the weighted rows. Table~\ref{tab:ablation_full} additionally retains the $\sigma=1$ result and content-only positional Gaussian controls.

\begin{table}[!htbp]
\small
\caption{Fixed-gain redistribution and normalization results, including the narrower Gaussian kernel and content-only controls. All intervention rows use $\lambda_w=2$, $\lambda_p=0.1$, and $\alpha=1$. Content-only Gaussian redistributes according to token-position distance and preserves content mass before weighting.}
\label{tab:ablation_full}
\begin{center}
\begin{tabular}{llrrr}
\toprule
Redistribution & Row norm. & SSCD $\downarrow$ & CLIP $\uparrow$ & ImageReward $\uparrow$\\
\midrule
None & No & 0.2052 & 0.2604 & $-0.3663$\\
\textbf{Gaussian (ours)}, $\sigma=1$ & No & 0.2200 & 0.2737 & $-0.2041$\\
\textbf{Gaussian (ours)}, $\sigma=4$ & No & 0.2088 & 0.2687 & $-0.2580$\\
Content Gaussian ($\sigma_c=1$, $\rho=1$) & No & 0.2224 & 0.2746 & $-0.2112$\\
Uniform (content-only, $\rho=0.5$) & No & 0.2279 & 0.2751 & $-0.2086$\\
\midrule
None & Yes & 0.3831 & 0.2913 & 0.0467\\
\textbf{Gaussian (ours)}, $\sigma=4$ & Yes & 0.3312 & 0.2793 & $-0.0231$\\
Content Gaussian ($\sigma_c=1$, $\rho=1$) & Yes & 0.3458 & 0.2851 & 0.0245\\
Uniform (content-only, $\rho=0.5$) & Yes & 0.3556 & 0.2875 & 0.0449\\
\midrule
Unmodified base & -- & 0.5465 & 0.3028 & 0.0594\\
\bottomrule
\end{tabular}
\end{center}

\vspace{-3pt}
\end{table}

\FloatBarrier
\section{Content reinforcement and the suppression boundary}
\label{app:reinforcement}
\begin{figure}[!htbp]
\centering
\includegraphics[width=0.62\linewidth]{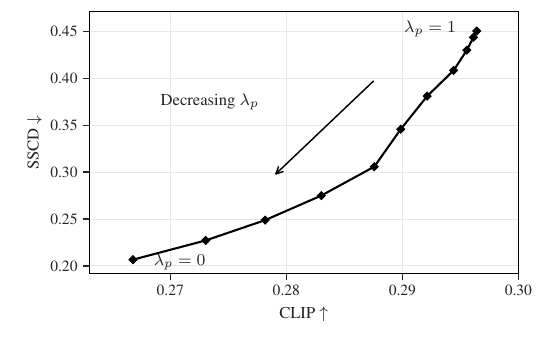}
\par\vspace{-3pt}
\caption{\textbf{Padding suppression trades alignment for lower similarity.} Decreasing $\lambda_p$ from 1 to 0 lowers both SSCD and CLIP in the SD v1.4 sweep. Markers follow the recorded settings; the line connects successive settings.}
\label{fig:padding_sweep}
\vspace{-3pt}
\end{figure}

\begin{figure}[!htbp]
\centering
\includegraphics[width=\linewidth]{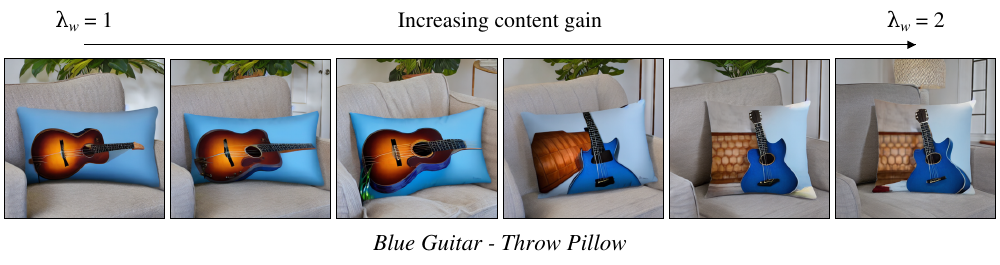}
\par\vspace{-3pt}
\caption{\textbf{Content reinforcement supports the requested attribute.} The complete six-output progression for \emph{Blue Guitar -- Throw Pillow}, ordered by increasing content gain. At higher gain, the guitar printed on the pillow is blue.}
\label{fig:gain_examples}
\vspace{-3pt}
\end{figure}

The content-gain and padding-gain sweeps use the SD v1.4 benchmark protocol with ten images per prompt. The content-gain sweep fixes the Gaussian width at $\sigma=4$. With the Gaussian intervention retained, the CLIP panel covers $\lambda_w$ from 1 to 2, whereas the SSCD panel extends to 3. CLIP increases over its recorded range; SSCD first rises before declining at larger gains. Figure~\ref{fig:reinforcement}(c) shows the endpoints at gains 1 and 2 and an intermediate output, with the guitar itself appearing blue at gain 2; Figure~\ref{fig:gain_examples} retains the full image progression. Figure~\ref{fig:padding_sweep} shows that decreasing $\lambda_p$ from 1 toward 0 lowers SSCD together with CLIP. This trade-off motivates pairing suppression with content reinforcement to control the alignment cost.

Figures~\ref{fig:reinforcement}(a,b) and~\ref{fig:padding_sweep} redraw the recorded sweeps from markers digitized from the original plots, connecting successive settings without curve fitting. Original plots, extracted coordinates, and calibration details are retained with the manuscript source.

The separate EOS-boundary experiment uses four images for each of the 500 prompts (112 VM and 388 TM), DDIM with 50 steps, CFG 7, and matched initial random draws. Both arms use probability-space Gaussian smoothing without row normalization, $\lambda_w=2$, $\lambda_p=0.1$, and a five-step redistribution window. The sole change at each width is the multiplier at the first EOS: 1 for our method and 0.1 for the EOS-suppressed variant. The base scores for this run are SSCD 0.5486, CLIP 0.299, and ImageReward $-0.108$. Table~\ref{tab:eos} reports all eight intervention settings.

\begin{table}[!htbp]
\small
\caption{Effect of EOS suppression within our method on 500 SD v1.4 prompts, with four images per prompt. Only the first EOS multiplier changes: 1 when preserved and 0.1 when suppressed.}
\label{tab:eos}
\begin{center}
\begin{tabular}{rrrrrrr}
\toprule
& \multicolumn{3}{c}{EOS preserved (Ours)} & \multicolumn{3}{c}{EOS suppressed}\\
\cmidrule(lr){2-4}\cmidrule(lr){5-7}
$\sigma$ & SSCD $\downarrow$ & ImageReward $\uparrow$ & CLIP $\uparrow$ & SSCD $\downarrow$ & ImageReward $\uparrow$ & CLIP $\uparrow$\\
\midrule
1 & 0.2332 & $-0.350$ & 0.272 & 0.2112 & $-0.509$ & 0.265\\
4 & 0.2082 & $-0.413$ & 0.262 & 0.1899 & $-0.570$ & 0.257\\
10 & 0.1656 & $-0.637$ & 0.249 & 0.1518 & $-0.762$ & 0.243\\
100 & 0.1453 & $-0.721$ & 0.243 & 0.1382 & $-0.823$ & 0.238\\
\bottomrule
\end{tabular}
\end{center}
\vspace{-3pt}
\end{table}

\begin{figure}[!htbp]
\centering
\includegraphics[width=\linewidth]{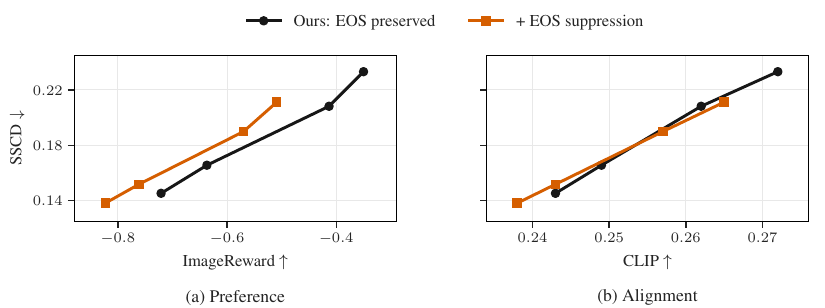}
\par\vspace{-3pt}
\caption{EOS-boundary trade-offs for CLIP and ImageReward. Lines connect the four evaluated widths per variant. Lower-right is preferable; SSCD uses the shared vertical scale.}
\label{fig:eos_complete}
\vspace{-3pt}
\end{figure}

The EOS boundary changes the suppression--quality trade-off. At each fixed width, extending attenuation to EOS lowers SSCD together with CLIP and ImageReward (Table~\ref{tab:eos}). Across widths, preserving EOS yields more favorable ImageReward--SSCD operating points, with a smaller separation in CLIP (Figure~\ref{fig:eos_complete}). The EOS-suppressed variant reaches a lower minimum SSCD (0.1382 versus 0.1453), accompanied by lower alignment and preference. Preserving EOS therefore retains prompt information and image preference while maintaining substantial suppression.

Bright Ending \citep{BrightEnding} uses EOS-token attention to obtain spatial memorization masks and incorporates them into detection and prompt optimization. Our boundary experiment isolates EOS weighting within our method; its relation to prior attention-based mitigation is discussed in Section~\ref{sec:related_work}.

\begin{figure}[!htbp]
\centering
\includegraphics[width=0.82\linewidth]{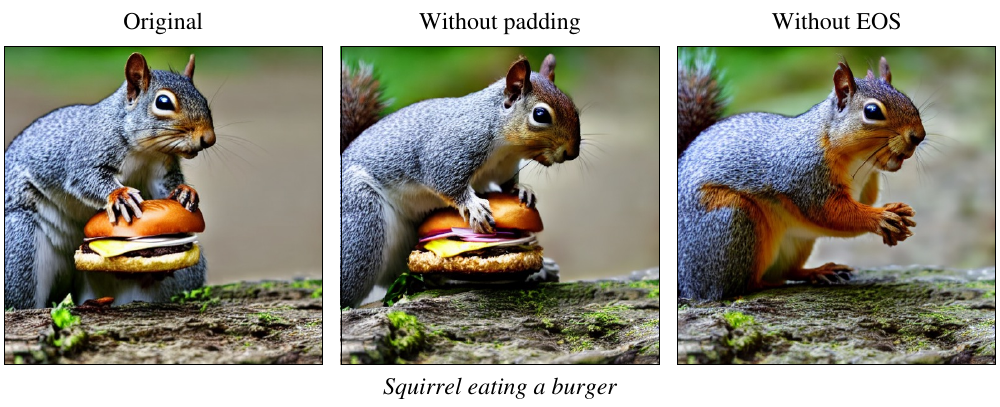}
\par\vspace{-3pt}
\caption{\textbf{EOS removal changes prompt content.} A token-removal example for \emph{Squirrel eating a burger}. Removing padding retains the depicted interaction more closely than removing EOS. This is a separate qualitative intervention from the gain-scaling experiment in Figure~\ref{fig:eos_complete}.}
\label{fig:eos_examples}
\vspace{-3pt}
\end{figure}

Figure~\ref{fig:eos_examples} complements the baby example in Figure~\ref{fig:eos} with a separate non-memorized prompt. Removing padding retains the squirrel--burger interaction, whereas removing EOS changes the depicted interaction. These examples motivate treating EOS separately from padding; the controlled gain-scaling comparison above evaluates that choice within our method.

\FloatBarrier
\section{Supplementary component comparisons}
\label{app:alternatives}
The global uniform alternative (label smoothing) modifies all $L$ token positions as $A'_i=(1-\beta)A_i+\beta\mathbf1/L$ before the common weighting steps. Temperature scaling modifies logits across the entire sequence. Both differ from Eq.~\eqref{eq:content}, which redistributes only existing content mass. The global smoothing comparison in Table~\ref{tab:global_smoothing} and the content-only controls therefore evaluate different redistribution domains.

\begin{table}[!htbp]\small
\caption{Global smoothing alternatives on SD v1.4 under the ten-image benchmark protocol. Global uniform smoothing differs from the content-only uniform variant in Table~\ref{tab:ablation}.}
\label{tab:global_smoothing}
\begin{center}
\begin{tabular}{llrr}
\toprule
Global smoothing alternative & Strength & SSCD $\downarrow$ & CLIP $\uparrow$\\
\midrule
Temperature & $\tau=1.5$ & 0.2927 & 0.2722\\
Temperature & $\tau=2.0$ & 0.1880 & 0.2237\\
Temperature & $\tau=2.5$ & 0.1143 & 0.1799\\
Global uniform & $\beta=0.01$ & 0.1847 & 0.2583\\
Global uniform & $\beta=0.05$ & 0.1965 & 0.2628\\
Global uniform & $\beta=0.10$ & 0.2212 & 0.2694\\
Gaussian & $\sigma=4$ & 0.2265 & 0.2731\\
Gaussian & $\sigma=10$ & 0.1838 & 0.2585\\
Gaussian & $\sigma=100$ & 0.1553 & 0.2484\\
\bottomrule
\end{tabular}
\end{center}

\vspace{-3pt}
\end{table}

\FloatBarrier
\section{Mechanism controls and interpretation}
\label{app:mechanism}
The direction--norm controls in Section~\ref{sec:controls} use the row-normalized content-positional setting (\texttt{abl\_position\_norm1}), with $\sigma_c=1$, $\rho=1$, $\alpha=1$, $\lambda_w=2$, and $\lambda_p=0.1$. Mixing is restricted to the content span with donor-normalized Gaussian positional weights, and the reweighted attention rows are normalized. This fixes attention mass, while the same-state swaps in Eq.~\eqref{eq:swap} separate guidance direction and magnitude. Table~\ref{tab:mechanism} reports the direction--norm decomposition of this setting. The main Gaussian operator uses reflected smoothing over all non-start positions and retains the reweighted row mass. The controls require two denoiser evaluations per step to obtain base and intervened guidance at the same latent. Sampling with our method retains the original number of denoiser evaluations.

Equation~\eqref{eq:swap} is evaluated separately at each current latent. In mixed precision, the realized direction/norm controls have maximum relative norm errors of approximately 2.02\% and 1.54\%, respectively; the corresponding means are 0.019\% and 0.029\%.

The mechanism analysis measures the conditional-guidance vector $\epsilon_c-\epsilon_u$ and its effect on generated-image similarity. The entropy diagnostic below measures the discrete distribution over token weights.

For Table~\ref{tab:mechanism}, paired prompt-bootstrap estimates give an SSCD difference from base of $-0.0477$ (95\% CI $[-0.0612,-0.0356]$) when only the guidance norm is changed. Retaining the intervention direction while restoring the base norm gives $-0.2032$ (95\% CI $[-0.2283,-0.1782]$). These intervals use the prompt-level resampling procedure in Appendix~\ref{app:implementation}.

In a separate early-step diagnostic, 500 memorized, 500 alignment-failure, and 500 normal prompts were evaluated over ten denoising steps, recording attention at spatial resolution $16\times16$. The cross-prompt correlation between normalized content entropy and guidance norm was $+0.204$ for memorized prompts, $-0.782$ for alignment-failure prompts, and $-0.434$ for normal prompts. After controlling prompt length, these correlations were $+0.245$, $-0.553$, and $-0.437$. The relationship is therefore population-dependent in this diagnostic. This motivates the controlled allocation--magnitude comparison.

\FloatBarrier
\section{Alternative SD operating point and detailed results}
\label{app:subgroups}
The moderate-strength comparison evaluates our Gaussian configuration with $\sigma=4$, $\alpha=0.625$, $\lambda_w=2$, and $\lambda_p=0.1$, against Content-uniform, a normalized content-only redistribution setting with $\rho=0.5$, $\alpha=1$, $\lambda_w=2^{1.5}$, and $\lambda_p=0.1^{1.5}$. The intervention strengths for this comparison were selected using SSCD measurements on a separate 160-prompt subset of the benchmark. The two operating points vary in redistribution domain, kernel, weighting, normalization, and mix. Their complete paired results appear in Table~\ref{tab:finalpair}. Content-uniform lowers SSCD and raises ImageReward with a small alignment cost. It reduces high-similarity batches by 4.80 percentage points.

\begin{table}[!htbp]\small
\caption{Moderate-strength comparison on 340 prompts and three seeds. Differences are Content-uniform minus Ours, with 95\% prompt-bootstrap CIs. High-similarity rates use 1,020 four-image batches per arm; the rate difference is in percentage points.}
\label{tab:finalpair}
\begin{center}
\begin{tabular}{lrrr}
\toprule Metric & Ours & Content-uniform & Difference [95\% CI]\\ \midrule
GT SSCD $\downarrow$ & 0.36327 & 0.34738 & $-0.01589$ [$ -0.02435, -0.00787 $]\\
CLIP $\uparrow$ & 0.28919 & 0.28701 & $-0.00219$ [$ -0.00381, -0.00058 $]\\
ImageReward $\uparrow$ & -0.07670 & -0.02627 & $+0.05043$ [$ +0.02581, +0.07574 $]\\
High-similarity batch (\%) $\downarrow$ & 50.20 & 45.39 & $-4.80$ [$ -7.06, -2.75 $]\\
\bottomrule\end{tabular}
\end{center}

\vspace{-3pt}
\end{table}

Component and mechanism analyses use seeds 509 and 811, with four images per prompt and seed. The moderate-strength comparison fixes both configurations before evaluation on three additional seeds, 1009, 1013, and 1019. In Table~\ref{tab:seed_breakdown}, the paired point estimates for SSCD, ImageReward, and the high-similarity event have the same direction across all three seeds.

\begin{table}[!htbp]\small
\caption{Paired differences (Content-uniform minus Ours) by recorded memorization type: TM, 264 prompts; VM, 76 prompts. Seeds are averaged within each prompt before bootstrapping. Event differences are percentage points.}
\label{tab:memorization_types}
\begin{center}
\begin{tabular}{llrr}
\toprule Type & Metric & Difference & 95\% CI\\ \midrule
TM & GT SSCD & $-0.00961$ & [$ -0.01628, -0.00335 $]\\
TM & CLIP & $-0.00330$ & [$ -0.00494, -0.00173 $]\\
TM & ImageReward & $+0.03629$ & [$ +0.01019, +0.06224 $]\\
TM & High-similarity batch (pp) & $-3.28$ & [$ -5.43, -1.39 $]\\
VM & GT SSCD & $-0.03773$ & [$ -0.06754, -0.00951 $]\\
VM & CLIP & $+0.00170$ & [$ -0.00276, +0.00643 $]\\
VM & ImageReward & $+0.09957$ & [$ +0.03288, +0.16914 $]\\
VM & High-similarity batch (pp) & $-10.09$ & [$ -16.67, -3.95 $]\\
\bottomrule\end{tabular}
\end{center}

\vspace{-3pt}
\end{table}
\begin{table}[!htbp]\small
\caption{Per-seed paired differences, Content-uniform minus Ours, over the same 340 prompts. SSCD differences include 95\% paired prompt-bootstrap CIs. $\Delta$event is the high-similarity batch-rate difference in percentage points.}
\label{tab:seed_breakdown}
\begin{center}
\begin{tabular}{rlrrr}
\toprule Seed & $\Delta$SSCD [95\% CI] & $\Delta$CLIP & $\Delta$ImageReward & $\Delta$event (pp)\\ \midrule
1009 & $-0.02906$ [$-0.03914, -0.01949$] & $-0.00276$ & $+0.04530$ & $-5.88$\\
1013 & $-0.00290$ [$-0.01406, +0.00780$] & $-0.00140$ & $+0.03608$ & $-4.12$\\
1019 & $-0.01573$ [$-0.02475, -0.00718$] & $-0.00240$ & $+0.06992$ & $-4.41$\\
\bottomrule\end{tabular}
\end{center}

\vspace{-3pt}
\end{table}

\begin{samepage}
In Table~\ref{tab:memorization_types}, the SSCD and ImageReward point estimates favor Content-uniform more strongly on VM than TM. The VM CLIP interval spans zero. For TM, the combined reward interval is positive, while CLIP decreases. These breakdowns show how the aggregate trade-off varies across memorization types.
\par\end{samepage}

\section{DeepFloyd IF protocol and results}
\label{app:if}
\paragraph{Fixed configurations transferred from SD.}
Our method uses the moderate SD configuration: global Gaussian redistribution with $\sigma=4$, $\alpha=0.625$, $\lambda_w=2$, $\lambda_p=0.1$, and no row normalization. Content-uniform uses $\rho=0.5$, $\alpha=1$, $\lambda_w=2^{1.5}$, $\lambda_p=0.1^{1.5}$, and row normalization. Each retains its SD-selected settings when adapted to IF's text-key block. Their SD comparison (Table~\ref{tab:finalpair}) favors Content-uniform in SSCD and ImageReward, with a small CLIP decrease. The IF experiment compares the transfer of these complete configurations, including their redistribution domains, kernels, gains, normalization, and intervention strengths.

\paragraph{Comparison with the alternative SD configuration.}
Table~\ref{tab:if_full} reports all three arms of the transfer experiment. Content-uniform remains close to the base template-match rate while retaining higher CLIP and ImageReward than our configuration. The relative suppression--quality behavior of these SD-selected configurations therefore changes on IF.

\begin{table}[!htbp]
\small
\caption{Complete comparison of SD-selected configurations on DeepFloyd IF. Both interventions retain their SD settings. All arms use the same prompts, seeds, and template metrics as Table~\ref{tab:if}.}
\label{tab:if_full}
\begin{center}
\begin{tabular}{lrrrr}
\toprule
IF setting & Template matches / 144 & Template SSCD $\downarrow$ & CLIP $\uparrow$ & ImageReward $\uparrow$\\
\midrule
Base & 56 (38.89\%) & 0.1985 & 0.3045 & 0.3988\\
Content-uniform & 54 (37.50\%) & 0.1939 & 0.2991 & 0.4556\\
Ours & 29 (20.14\%) & 0.1528 & 0.2842 & 0.1058\\
\bottomrule
\end{tabular}
\end{center}
\end{table}

\paragraph{Prompt and reference selection.}
We use the published template dataset\footnote{\url{https://huggingface.co/datasets/fraisdufour/templates-verbs}} at revision \texttt{213a14db}. Its IF prompt table, \nolinkurl{deep_if_sdv1_wb_groundtruth.parquet}, contains the captions and historical annotations. A URL join links 160 TV prompts to the supplied references. Shared reference identifiers induce 11 connected groups. With selection seed 9162026, prompts are shuffled within each group and selected round-robin to reach 48 TV prompts; 16 historical N prompts are sampled separately. Two MV rows are outside the chosen evaluation scope. The transfer evaluation therefore targets template memorization. All selected TV prompts are retained in the evaluation regardless of whether their baseline outputs satisfy the template-match criterion.

TV prompts use seeds 0, 3, and 1301; historical N prompts use seed 1301. Stage I uses \texttt{IF-I-XL-v1.0} and Stage II \texttt{IF-II-L-v1.0}. Both use 50 steps, with guidance scales 7 and 4, respectively; Stage II noise level is 250. The Stage II random seed is the Stage I seed plus 1,000,003. T5 embeddings are computed using float32 activations from the downloaded fp16 checkpoint and cached in half precision. Token length is fixed at 77, and caption cleaning is disabled. All arms use the same attention backend and pretrained safety-checking pipeline; no generated images were excluded by tokenization or safety filtering.

For generated RGB image $x$ and reference image/mask $(r_j,m_j)$, resized to $256\times256$ and scaled to $[0,1]$, define
\begin{equation}
 S_j(x)=\frac{\sum_{u,v,c}[(r_{j,uvc}-x_{uvc})m_{j,uvc}]^2}
 {\operatorname{mean}(m_j)},\quad
 E_{\rm IF}(x)=\mathbf1\{\min_{1\leq j\leq57} S_j(x)<2500\}.
\end{equation}
The primary event searches all 57 reference pairs and measures the frequency of outputs satisfying the template-match criterion. Mean continuous masked error summarizes average discrepancy across generated outputs, including those outside the match region. Secondary template SSCD uses the maximum over the references assigned to that prompt. Restricting the threshold event to assigned references instead yields 30/144, 17/144, and 29/144 matches for base, our method, and the Content-uniform alternative, respectively; the qualitative conclusion is unchanged. The reduction in template-match frequency occurs without a clear change in mean continuous masked error.

For seeds 0, 3, and 1301, our method produces 8/48, 10/48, and 11/48 template matches, compared with 18/48, 17/48, and 21/48 for the base model. Content-uniform gives 18/48, 19/48, and 17/48. Template-match frequency is therefore lower for our transferred configuration at each evaluated seed.

For paired IF CIs, we first average the paired seed differences within each prompt, then resample the 11 reference groups with replacement 10,000 times (RNG seed 20260917). All prompts in each sampled group are retained, and the resulting mean is weighted by the number of sampled prompts. The template-match difference from base is $-18.75$ percentage points (95\% CI $[-34.07,-6.00]$) for our method and $-1.39$ points (95\% CI $[-10.32,5.69]$) for the Content-uniform alternative. The ImageReward difference from base is $-0.2930$ [$-0.4869,-0.1007$] for our method and $+0.0568$ [$-0.0400,0.1548$] for the Content-uniform alternative. The corresponding CLIP differences are $-0.02033$ [$-0.03001,-0.01015$] and $-0.00547$ [$-0.01181,0.00378$].

The 16 historical N prompts provide a supplementary quality check. No arm matched a template. CLIP/ImageReward means were 0.3150/0.5902 for base, 0.3047/0.3010 for our method, and 0.3073/0.5275 for the Content-uniform alternative.

\FloatBarrier
\section{Detector-gated pipeline}
\label{app:pipeline}
We use the initial-step norm $\|\epsilon_c-\epsilon_u\|$ and calibrate thresholds on a set of 500 normal prompts. The detector AUROC is 0.902 against normal prompts and 0.949 against the alignment-failure set in the recorded detection analysis. Detection recall and precision use measured baseline similarity ($\mathrm{SSCD}>0.5$) as the positive criterion, rather than treating every historical prompt label as a positive event. Table~\ref{tab:detector_sweep} reports the threshold sweep. At threshold 1.8135, recall and precision are approximately 0.55 and 0.87. This differs from the fraction of the entire 500-prompt historical memorized set that is flagged (57.8\%).

\begin{table}[!htbp]\small
\caption{Detector threshold sweep with our full-strength configuration. Flag rates are empirical measurements on the stated calibration and evaluation sets.}
\label{tab:detector_sweep}
\begin{center}
\begin{tabular}{rrrrr}
\toprule
Threshold & Normal flagged (\%) & Memorized flagged (\%) & Pipeline SSCD & Normal ImageReward\\
\midrule
1.429 & 20.0 & 80.8 & 0.2783 & 0.2345\\
1.814 & 5.0 & 57.8 & 0.3475 & 0.2302\\
2.104 & 1.0 & 51.8 & 0.3622 & 0.2340\\
3.736 & 0.0 & 32.2 & 0.4078 & 0.2366\\
\bottomrule
\end{tabular}
\end{center}

\vspace{-3pt}
\end{table}
At the 5\% operating point, 25 of 500 normal prompts and 289 of 500 historically memorized prompts are flagged; the remaining 211 memorized prompts are not treated. Average CLIP changes are $-0.0002$ over the full normal set and $-0.0039$ over the 25 flagged prompts. The corresponding reward changes are $-0.0065$ and $-0.1291$. Population-average costs depend on flag prevalence. The pipeline uses our full-strength Gaussian configuration throughout.

\FloatBarrier
\section{Complete baseline results}
\label{app:legacy}
Tables~\ref{tab:baseline_sd14} and~\ref{tab:baseline_sd20} report all evaluated operating points underlying Figure~\ref{fig:legacy}, using ten images per prompt. The number of settings varies by method. For AIN, the implementation identifiers \texttt{aug\_latent} and \texttt{init\_noise\_bw} denote the per-sample and batch-wise variants, respectively. Learning rates, optimization iterations, targets, and batch parameters are given for reproducibility.
\begin{table}[!htbp]\scriptsize
\caption{Complete SD v1.4 aggregate results used in Figure~\ref{fig:legacy}. AIN parameters follow the implementation notation. Values are rounded to four decimal places.}
\label{tab:baseline_sd14}
\begin{center}
\begin{tabular}{llrrr}
\toprule Method & Recorded setting & SSCD & CLIP & ImageReward\\ \midrule
Base & -- & 0.5439 & 0.3012 & -0.0140\\
Ours & $\sigma=1$ & 0.2331 & 0.2773 & -0.1868\\
Ours & $\sigma=4$ & 0.2265 & 0.2731 & -0.2374\\
Ours & $\sigma=10$ & 0.1838 & 0.2585 & -0.3779\\
Ours & $\sigma=100$ & 0.1553 & 0.2484 & -0.5188\\
RTA & tokens=1 & 0.4794 & 0.2903 & -0.1013\\
RTA & tokens=2 & 0.4171 & 0.2778 & -0.1581\\
RTA & tokens=4 & 0.3402 & 0.2608 & -0.3532\\
RTA & tokens=6 & 0.2728 & 0.2405 & -0.5496\\
GPA & target=3 & 0.1325 & 0.2372 & -0.6466\\
GPA & target=4 & 0.1994 & 0.2545 & -0.4430\\
GPA & target=5 & 0.2527 & 0.2644 & -0.3338\\
GPA & target=6 & 0.3117 & 0.2733 & -0.2510\\
CS & $C=1.15$ & 0.3094 & 0.2732 & -0.1236\\
CS & $C=1.2$ & 0.2935 & 0.2696 & -0.1666\\
CS & $C=1.25$ & 0.2779 & 0.2661 & -0.2115\\
CS & $C=1.3$ & 0.2662 & 0.2632 & -0.2506\\
DTP & scale=1 & 0.2396 & 0.2409 & -0.6639\\
DTP & scale=3 & 0.2136 & 0.2403 & -0.6772\\
DTP & scale=5 & 0.1908 & 0.2315 & -0.7660\\
DTP & scale=7 & 0.1711 & 0.2202 & -0.8884\\
AIN (batch) & rho=50, gamma=0.7,iters=2,CFGstep=4 & 0.2642 & 0.2603 & -0.4527\\
AIN (sample) & lr=0.01, target=0.7, iters=1000 & 0.1731 & 0.2505 & -0.5552\\
AIN (sample) & lr=0.01, target=0.9, iters=1000 & 0.2265 & 0.2647 & -0.3522\\
AIN (sample) & lr=0.01, target=1.1, iters=1000 & 0.2770 & 0.2743 & -0.2556\\
AIN (sample) & lr=0.01, target=1.3, iters=1000 & 0.3220 & 0.2810 & -0.1847\\
AIN (sample) & lr=0.01, target=1.5, iters=1000 & 0.3593 & 0.2850 & -0.1414\\
\bottomrule\end{tabular}
\end{center}

\vspace{-3pt}
\end{table}
\begin{table}[!htbp]\scriptsize
\caption{Complete SD v2.0 aggregate results used in Figure~\ref{fig:legacy}. AIN parameters follow the implementation notation. Values are rounded to four decimal places.}
\label{tab:baseline_sd20}
\begin{center}
\begin{tabular}{llrrr}
\toprule Method & Recorded setting & SSCD & CLIP & ImageReward\\ \midrule
Base & -- & 0.2949 & 0.2640 & -0.4280\\
Ours & $\sigma=1$ & 0.2090 & 0.2579 & -0.5482\\
Ours & $\sigma=4$ & 0.1654 & 0.2208 & -1.0782\\
Ours & $\sigma=10$ & 0.1638 & 0.2201 & -1.1037\\
Ours & $\sigma=100$ & 0.1617 & 0.2199 & -1.1275\\
RTA & tokens=1 & 0.2882 & 0.2620 & -0.4536\\
RTA & tokens=2 & 0.2836 & 0.2587 & -0.5151\\
RTA & tokens=4 & 0.2663 & 0.2517 & -0.5647\\
RTA & tokens=6 & 0.2319 & 0.2356 & -0.8017\\
RTA & tokens=8 & 0.2004 & 0.2164 & -0.9984\\
GPA & target=40 & 0.1550 & 0.1722 & -1.5205\\
GPA & target=50 & 0.1893 & 0.2052 & -1.2022\\
GPA & target=60 & 0.2145 & 0.2261 & -0.9690\\
GPA & target=70 & 0.2456 & 0.2457 & -0.7365\\
CS & $C=1.2$ & 0.2574 & 0.2508 & -0.6474\\
CS & $C=1.25$ & 0.2485 & 0.2475 & -0.6975\\
CS & $C=1.3$ & 0.2407 & 0.2417 & -0.7608\\
CS & $C=1.35$ & 0.2338 & 0.2366 & -0.8394\\
DTP & scale=1 & 0.1570 & 0.1733 & -1.4934\\
DTP & scale=3 & 0.1354 & 0.1533 & -1.6944\\
DTP & scale=5 & 0.1162 & 0.1277 & -1.9013\\
DTP & scale=7 & 0.1051 & 0.1108 & -2.0174\\
AIN (sample) & lr=0.1, target=20, iters=10 & 0.2740 & 0.2631 & -0.4822\\
AIN (sample) & lr=0.1, target=15, iters=10 & 0.2530 & 0.2576 & -0.5885\\
AIN (sample) & lr=0.1, target=10, iters=10 & 0.2254 & 0.2479 & -0.7735\\
AIN (sample) & lr=0.1, target=8, iters=50 & 0.1984 & 0.2340 & -0.9445\\
AIN (sample) & lr=0.1, target=5, iters=50 & 0.1712 & 0.2154 & -1.2437\\
AIN (batch) & rho=30, gamma=0.7,iters=4,CFGstep=4 & 0.1976 & 0.2335 & -0.9549\\
\bottomrule\end{tabular}
\end{center}

\vspace{-3pt}
\end{table}

\clearpage
\section{Qualitative behavior}
\label{app:qualitative}
The following examples extend the qualitative evaluation with semantic prompt edits and the complete content-gain progression.

\begin{figure}[!htbp]
\centering
\includegraphics[width=\linewidth]{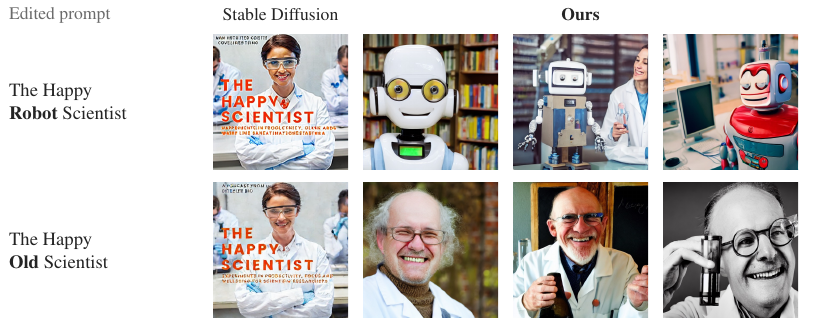}
\par\vspace{-3pt}
\caption{\textbf{Response to semantic prompt edits.} Adding \emph{Robot} or \emph{Old} changes the subject in our outputs, whereas the shown SD outputs remain close to a common book-cover template. Each row shows one SD output and three outputs from our method for the same edited prompt.}
\label{fig:semantic_edits}
\vspace{-3pt}
\end{figure}

\end{document}